\documentclass[11pt]{article}

\usepackage{acl}

\usepackage{times}
\usepackage{latexsym}

\usepackage[T1]{fontenc}
\usepackage[utf8]{inputenc}

\usepackage{microtype}

\usepackage{inconsolata}
\usepackage{booktabs}
\usepackage{multirow}
\usepackage{tabularx}
\usepackage{float}
\usepackage{graphicx}
\usepackage{subcaption}

\usepackage{graphicx}
\usepackage{colortbl}

\usepackage[table]{xcolor}
\usepackage{pifont}

\definecolor{g1}{RGB}{218,237,223}  % Barely lighter
\definecolor{g2}{RGB}{186,224,196}
\definecolor{g3}{RGB}{148,208,165}
\definecolor{g4}{RGB}{105,191,127}
\definecolor{g5}{RGB}{67,172,93}
\definecolor{g6}{RGB}{44,150,76}
\definecolor{g7}{RGB}{21,127,60}     % Very dark green
\newcommand{\heatcell}[2]{\cellcolor{#1}#2}

\usepackage{svg}
\usepackage{amsmath}
\usepackage{amssymb}
\usepackage{mathtools}
\usepackage{amsthm}
\usepackage[most]{tcolorbox}
\usepackage[small,compact]{titlesec}
\title{Compliance vs. Sensibility: On the Reasoning Controllability in Large Language Models}

\author{Xingwei Tan$^{\;\alpha}$ \quad Marco Valentino$^{\;\alpha}$ \quad  Mahmud Elahi Akhter$^{\;\beta}$ \\ \textbf{Yuxiang Zhou}$^{\;\beta\;\omega}$ \textbf{Maria Liakata}$^{\;\beta\;\gamma}$ \quad  \textbf{Nikolaos Aletras}$^{\;\alpha}$ \\
    $^\alpha$University of Sheffield \quad
    $^\beta$Queen Mary University of London \\
    $^\gamma$The Alan Turing Institute \quad 
    $^\omega$Ufonia\\
    \texttt{\{xingwei.tan,m.valentino,n.aletras\}@sheffield.ac.uk}\\
    \texttt{\{m.akhter,yuxiang.zhou,m.liakata\}@qmul.ac.uk}}

\begin{document}
\maketitle
\begin{abstract}
Large Language Models (LLMs) acquire reasoning capabilities through shared inference patterns in pre-training data, which are further elicited via Chain-of-Thought (CoT). However, whether fundamental reasoning patterns, such as induction, deduction, and abduction, can be decoupled from specific problem instances remains a critical challenge for model controllability. In this paper, we present the first systematic investigation of this problem through the lens of \emph{reasoning conflicts}, an explicit tension between parametric and contextual information induced by mandating logical schemata that deviate from those expected for a target task. Our evaluation reveals that LLMs consistently prioritize \textit{sensibility} over \textit{compliance}, favoring task-appropriate reasoning patterns despite conflicting instructions. We further demonstrate that reasoning conflicts are internally detectable, as confidence scores drop during conflicting episodes. 
Probing reveals instruction decodability even without compliance, while CKA identifies geometric differences across models and spans. Leveraging these insights, we steer models towards compliance, increasing instruction following by up to 29\%. Overall, our findings establish that while LLM reasoning is anchored to concrete instances, active mechanistic interventions can effectively decouple logical schemata from data, offering a path toward improved controllability, faithfulness, and generalizability.\footnote{The experimental code can be found in: \url{https://github.com/Xingwei-Tan/compliance_sensibility}}
\end{abstract}

\section{Introduction}

Recent work has shown that Large Language Models (LLMs) acquire fundamental reasoning capabilities by leveraging shared inference patterns and procedural knowledge within their pre-training data \citep{ruis2025procedural}. These capabilities, further elicited via Chain-of-Thought (CoT) prompting, offer remarkable performance on logical reasoning benchmarks \citep{wei2022chain, NEURIPS2022_8bb0d291,sprague2025to}. 
By prompting the model to generate a step-by-step sequence of intermediate thoughts, CoT enables LLMs to decompose complex problems and justify their final answer.

\begin{figure*}[ht]
    \centering
    \includegraphics[width=0.85\textwidth]{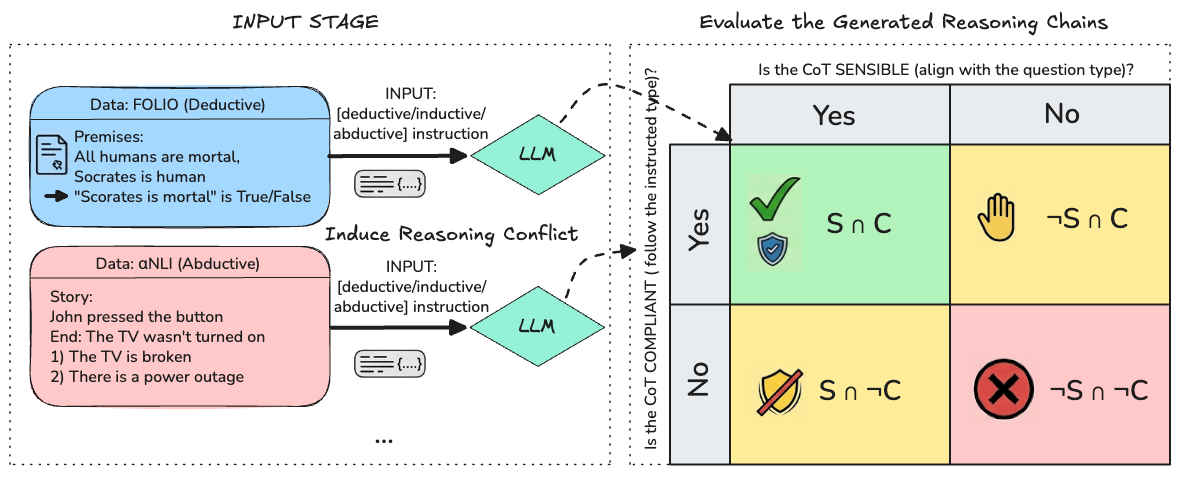}
    \caption{Reasoning instructions are used to induce reasoning conflicts. Then, we evaluate the sensibility and compliance of the responses.}
    \label{fig:intro}
    \vspace{-8pt}
\end{figure*}

However, it is unclear whether the inference patterns employed during CoT derive from generalizable, disentangled reasoning capabilities or remain intrinsically tied to the specific tasks and instances observed during pre-training. 
Standard CoT typically leaves the choice of the actual reasoning mechanism entirely up to the model. The LLM acts as a black box in deciding whether to deploy deductive, inductive, or abductive reasoning \citep{walton2001abductive}. 
If it naturally selects the correct reasoning strategy for the task, this autonomy is highly effective. 
But the lack of explicit structural controllability raises robustness and faithfulness issues. If a user accidentally applies the wrong prompt template, or malevolently injects an adversarial instruction forcing a suboptimal logical path, it is not clear how models react. 
When a human is handed a problem and told to use an inappropriate reasoning schema such as deduction for an ambiguous abductive mystery, they rely on meta-reasoning and cognitive flexibility to handle the conflict~\citep{ackerman2017meta, diamond2013executive}. 
Humans continuously monitor their own reasoning processes~\citep{nelson1990metamemory,lemaire1995strategy}, hence they might recognize the mismatch and switch to the task-appropriate reasoning, or attempt to follow the request, realize it is failing, then revise it mid-thought. 
It is currently unknown how LLMs navigate this exact tension, leading to a fundamental, yet underexplored, research question:~\textit{How robust and controllable is LLM reasoning when forced into a conflict between instruction compliance and logical sensibility?}

To answer this question, we first prepare a testbed consisting of logical reasoning questions that only require one of the fundamental reasoning types: deduction, induction, and abduction (Table \ref{fig:intro}). 
Then, we evaluate how LLMs react to instructions that align or conflict with the task-appropriate reasoning.
Our experiments reveal that a reasoning conflict between compliance (i.e., following the instruction) and sensibility (i.e., task-appropriate reasoning) indeed exists: models often face a tension between a user's reasoning instructions and their own internal priors.
To examine this gap, we combine linear probing with representation-similarity analysis: instructed reasoning types remain decodable in non-compliant responses, while the geometric effects of instruction changes vary across models and spans.
We show that the decision to comply or diverge is an active process that begins mid-computation. 
Finally, we use Contrastive Activation Addition (CAA) \citep{rimsky-etal-2024-steering} to showcase that reasoning compliance can be augmented during inference. 
Our contributions are as follows:
\begin{itemize}
    \item \textbf{Comprehensive Systematic Evaluation:} We conduct the first rigorous assessment of reasoning conflicts across three open-source model families and two frontier LLMs on four logical reasoning datasets, presenting a comprehensive evaluation of how LLMs direct their reasoning on diverse logical questions (\S\ref{sec:method}, \S\ref{sec:exp_setup}, and \S\ref{sec:exp_1}).
    \item \textbf{Mechanistic Analysis of Reasoning Prompting:} We provide the first in-depth investigation into the interplay between a user’s reasoning instructions and an LLMs' own internal priors for reasoning. Furthermore, we quantify instruction compliance under varying reasoning types and instruction settings (\S\ref{sec:mechanistic}).
    \item \textbf{Enhancing Reasoning Compliance via Steering:} We explore the efficacy of steering mechanisms in augmenting model compliance which shows steering can enhance the reasoning type compliance by up to $29\%$ (\S\ref{sec:steering}). 
\end{itemize}

\section{Related Work}

\subsection{Fundamental Logical Reasoning in LLMs}
Although LLMs have demonstrated remarkable reasoning capabilities on math, such as GSM8K \citep{cobbe2021trainingverifierssolvemath}, they still face various issues in logical reasoning or soft reasoning \citep{sprague2025to,liu2025logicalreasoninglargelanguage,sun2025languagemodelsfollowoccams}.
Recent work has shown that there is no one-size-fits-all reasoning strategy for LLMs, and they need to reason adaptively based on the context \citep{zheng-etal-2025-logidynamics}.
\citet{xu2025large} evaluate the fundamental reasoning paradigms and reveal a substantial performance gap: models frequently fail to identify the correct reasoning direction, particularly in abductive and mixed-form tasks.
\citet{liu2025evaluatinglogicalreasoningabilities} find similar gaps persist in inherently reasoning models such as DeepSeek-R1 \citep{guo2025deepseek} and Qwen3 \citep{yang2025qwen3technicalreport}.
This struggle is further evidenced in real-world claim verification, where even frontier LLMs fail on abductive reasoning questions \citep{dougrez-lewis-etal-2025-assessing}.
Recent efforts examine the structural interplay between the reasoning types, and investigate how different fundamental reasoning paradigms induce generalization or enhance LLMs' reasoning \citep{hu2025ahasystematicmetaabilitiesalignment,cao2026fundamentalreasoningparadigmsinduce}.

\subsection{Chain-of-Thought Faithfulness}
A growing body of evidence suggests that CoT frequently fails to faithfully describe the model's internal reasoning process \citep{turpin2023languagemodelsdontsay, chen2025reasoningmodelsdontsay}.
This discrepancy is often characterized as a lack of causal dependence, where the generated rationale serves as a post-hoc justification rather than a functional roadmap to the final answer \citep{lanham2023measuringfaithfulnesschainofthoughtreasoning, siegel-etal-2024-probabilities}.
Recent studies on soft-reasoning tasks indicate that while CoT may be unfaithful, it can still impose an active guidance effect on LLMs' output \citep{lewis-lim-etal-2025-analysing}, even if that guidance does not follow a direct causal link \citep{tutek2025measuringfaithfulnesschainsthoughtunlearning}.

Our work is related to injecting misleading cues to detect unfaithfulness \citep{turpin2023languagemodelsdontsay, chua2025deepseekr1reasoningmodels}, but we pivot the focus toward \textit{reasoning compliance}. While prior research has investigated whether models verbalize known causal features \citep{bao-etal-2025-likely}, it remains unclear how the verbalization relates to a model's ability to adhere to a mandated logical schema. By exploring how CoT influences predictions when a model either acknowledges or unfaithfully omits an explicit structural cue, we move beyond simple error-testing \citep{lanham2023measuringfaithfulnesschainofthoughtreasoning} to investigate the fundamental reasoning conflicts between user instructions and internal model priors.
This is particularly important for high-stake domains, such as the legal domain, where users would like full transparency on the reasoning process and full control on the reasoning schema \cite{kelsall2026rapid}.

\subsection{Mechanistic Interpretation and Model Steering for Logical Reasoning}
Recent mechanistic interpretation studies suggest that instruction-following and task-specific logic are often localized within residual stream activations, typically maturing in the middle-to-late layers of the transformer architecture \citep{belrose2025elicitinglatentpredictionstransformers, nanda2023progress}.
To bridge the gap between understanding these representations and controlling them, model editing is a lightweight and flexible alternative to fine-tuning \citep{hughes-etal-2026-patch}.  
While early applications focused on mitigating toxicity or enhancing safety, recent work applies these methods to reasoning. For instance, \citet{valentino2026mitigatingcontenteffectsreasoning} adopt Conditional Activation Steering \citep{lee2025programming} to mitigate content effects.
While prior steering research often targets binary attributes, we test CAA \citep{rimsky-etal-2024-steering} to the more abstract domain of reasoning compliance by sampling responses on our reasoning conflict testbed.

\section{Methodology}
\label{sec:method}

\subsection{Problem Formulation}
\label{sec:definiton}
We aim to measure how \textit{controllable} is LLM reasoning when forced into a conflict between \textit{instruction compliance} and \textit{logical sensibility} in logical reasoning problems.
We denote $\mathbf{q}$ as a question that requires logical reasoning and $\mathbf{a}$ as its correct answer.
We assume $\mathbf{q}$ is associated with an a priori known suitable reasoning type $\mathbf{t}$, which is predetermined by the dataset's creators. 
$\mathbf{t}$ can be deductive, inductive, or abductive reasoning.
An instruction $\mathbf{g_t}$ (defined in \S\ref{prompt_definition}) is formulated that includes a mandated fundamental reasoning type $\mathbf{t}'$ for guiding the LLMs to reason and answer the question.
These instructions are presented as part of the in-context prompt that will be fed into the LLMs.
The answer generation process is formulated as:
\begin{equation}
    P(\hat{r},\hat{a})=LLM(q,g_{t'}), t'\in\{\text{De., In., Ab.}\}
\end{equation}
where $\hat{a}$ is the generated final answer, and $\hat{\mathbf{r}}$ is the reasoning chain generated by CoT. Finally, we automatically infer the reasoning type $\hat{\mathbf{t}}$ used in $\hat{\mathbf{r}}$, and compare it with the mandated and suitable reasoning types, $\mathbf{t}'$ and $\mathbf{t}$ respectively.
When $\hat{\mathbf{t}}=\mathbf{t}'$, the model is \textbf{compliant} (C).
When $\hat{\mathbf{t}}=\mathbf{t}$, the model is \textbf{sensible} (S).
In some of the experiment settings, we intentionally prompt with $g_{t'},t'\neq t$ to induce a reasoning conflict.
Although part of a long-form generation may contains a secondary reasoning type, our formulation operates under a \textit{Dominant Reasoning Mode} assumption to build a controlled testbed for analyzing our research question.
We define $\hat{\mathbf{t}}$ as the most prominent reasoning type that contributes directly to the final answer.
All questions that we include in the experiments are specifically designed to be answered with only one reasoning type.

\subsection{Prompting for Reasoning Controllability}
\label{prompt_definition}
As prompting is the most accessible way a user would interact with an LLM, we first test reasoning controllability in-context via prompting. We want to quantify how well LLMs understand fundamental reasoning types, and how well they can perform them based on basic instructions. For that purpose, we construct zero-shot prompts $\mathbf{g_t}$ to control reasoning by showing LLMs the high level definition of the available reasoning types, and the general steps to perform them. Table~\ref{tab:prompts} shows the prompts for deductive, inductive and abductive reasoning.

\begin{table}[ht] 
\scriptsize 
\centering
\begin{tabularx}{0.95\columnwidth}{l X}
\toprule
\textbf{Type} & \textbf{Prompt} \\
\midrule
\textbf{Deductive} & You are a logician tasked with performing \textbf{deductive} reasoning. You are given a general rule and specific observations. Your task is to apply the general rule to the observations to derive a logically certain conclusion. Provide a detailed reasoning process leading to your conclusion. \\
\midrule
\textbf{Inductive} & You are a logician tasked with performing \textbf{inductive} reasoning. You are given a set of observations. Your task is to infer the most probable general rule that explains all observations. Then, use the inferred rule to make predictions about new observations. Provide a detailed reasoning process leading to your conclusion. \\
\midrule
\textbf{Abductive} & You are a logician tasked with performing \textbf{abductive} reasoning. You are given a general rule and an observation. Your task is to generate the most probable and simplest hypothesis that, if true, would logically explain all the observations provided. Provide a detailed reasoning process leading to your conclusion. \\
\bottomrule
\end{tabularx}
\caption{Prompts for controlling LLMs' reasoning.}
\label{tab:prompts}
\vspace{-8pt}
\end{table}

\subsection{Automated Reasoning Type Classification}
\label{sec:automated_reasoning_classification}

Given a generated CoT $\hat{\mathbf{r}}$, we classify the reasoning type $\mathbf{t'}$ employed by the model as deductive, inductive or abductive. Since this is challenging and expensive for manual annotation, as models often produce verbose and structurally ambiguous outputs, we use the LLM-as-a-Judge framework \citep{10.5555/3666122.3668142}.

We present the generated CoT to a frontier model using a zero-shot instruction: \textit{What type of reasoning is the following process?}
To formalize the judge's output, we employ a rule-based parser that maps responses into four categories: Deductive, Inductive, Abductive, or Direct (no intermediate reasoning).
The Cohen's $\kappa$ agreement between humans and \textsc{GPT-5.1} is $0.84$.
Notably, we find that frontier models are highly sensitive to lexical mimicking, where an LLM superficially echoes terms from the reasoning instructions (e.g., ``logically certain'' for deduction or ``most probable'' for abduction) without executing the corresponding logical operations (see \S\ref{sec: case_study}). 
The details of judge verification can be found in Appendix \ref{app:judge_verification}, where we also show the confusion matrices of the labels.

Finally, based on the comparison between the mandated and generated CoT reasoning type, each response is further categorized into the following four categories:
\begin{itemize}
    \item Sensible and Compliant ($S \cap C$): The model adopts the instructed reasoning type, which happens to be the task-appropriate one.
    \item Sensible but Non-Compliant ($S \cap \neg C$): The model ignores the reasoning instruction and defaults to the task-appropriate type.
    \item Compliant but Non-Sensible $(\neg S \cap C)$: The model follows the instruction even when it is logically ill-suited for the task (e.g., using induction for a deductive logic puzzle).
    \item Neither Sensible nor Compliant $(\neg S \cap \neg C)$: The model uses an unrelated reasoning type. 
\end{itemize}

\subsection{Steering for Reasoning Controllability}
We adopt model steering to enhance the controllability of LLMs by guiding them towards compliance with the reasoning instructions $\mathbf{t'}$.
We use CAA \citep{rimsky-etal-2024-steering} to identify the parameters of the LLM that are responsible for reasoning. CAA modifies internal activations during inference without updating model parameters. It effectively enables fine-grained control of sycophancy, refusal, and instruction-following behaviors \citep{stolfo2025improving}.
Given multiple pairs of contrastive responses $(r_p, r_n)$ of the question $q$ (target behavior and the opposite), a steering vector $\mathbf{v}$ is generated by computing the Mean Difference of the residual stream activations at a specific layer.
\begin{equation}
\small
    \mathbf{v}_{MD} = \frac{1}{|D|} \sum_{(q, r_p, r_n) \in D} \left( h_L(q, r_p) - h_L(q, r_n) \right),
\end{equation}
where $h_L(\cdot)$ is the activation at layer $L$ and $D$ the training dataset.
During inference, the steering vector is added to the corresponding layer at every token position.
A multiplier (coefficient) $\mu$ controls the intensity and direction of the effect.

To obtain a steering vector of instruction compliance, we collect compliant and non-compliant responses as contrastive pairs by sampling reasoning responses with the prompts in Table \ref{tab:prompts} on the training datasets with $3$ random seeds.
Then, the LLM judge introduced in \S\ref{sec:automated_reasoning_classification} determines whether the responses are compliant.
To make sure that only the reasoning compliance is contrasted, we marginalize the effects of final answer correctness and instructed type by only pairing those responses with the same correctness and instructed type.

\section{Experimental Setup}
\label{sec:exp_setup}

\subsection{Models}
We evaluate two frontier proprietary models: \textsc{GPT-5.1} \citep{openai2025gpt5card} and \textsc{Gemini3-flash} \citep{gemini2025}.
Three famil of state-of-the-art open-weight LLMs are also included in our experiments: \textsc{Olmo}~\citep{olmo2025olmo3}, \textsc{Qwen}, and \textsc{Llama}.
We compare smaller versions (7B/8B), larger version (32B/70B), instruct version, and thinking version of these open-weight LLMs.
For \textsc{Qwen}, we include the results with thinking mode on and off.

\subsection{Data}
We evaluate the LLMs across four datasets specifically selected to represent the fundamental reasoning types: deduction, induction, and abduction.

\textbf{FOLIO} (Deduction): A first-order logic dataset containing textual premises and claims \citep{han-etal-2024-folio}. Models must determine the truth value of a claim based strictly on the provided logical constraints. We utilize \textsc{FOLIO} as a benchmark for deductive reasoning, where conclusions must follow necessarily from established premises.

\textbf{SPR} (Induction): The \textit{Sequence Pattern Recognition} (SPR) task requires models to predict the subsequent element in a numerical sequence governed by recurring arithmetic operations \citep{hu2025ahasystematicmetaabilitiesalignment}. This task serves as a proxy for inductive reasoning, requiring the model to generalize a latent rule from observed patterns.

\textbf{$\boldsymbol{\alpha}$NLI} (Abduction): The \textit{Abductive Natural Language Inference} ($\alpha NLI$) dataset presents a story's beginning and ending, requiring the model to select the most plausible intermediate hypothesis from two options \citep{zhao-etal-2023-abductive}. This represents abductive reasoning, which is about inference to the best explanation.

\textbf{RECV} (Deduction+Abduction): \textit{Reasoning in Evidence-based Claim Verification} (\textsc{RECV}) dataset consists of real-world claim verification tasks sourced from Wikipedia and Twitter \citep{dougrez-lewis-etal-2025-assessing}. Unlike the other benchmarks, every \textsc{RECV} question is associated with a human-annotated reasoning type, providing a diverse testbed for reasoning compliance.

\subsection{Evaluation}
\textbf{Task Accuracy.}
We extract the final answer from the generated output via a rule-based parser that matches the special tags we put in the prompt template (Appendix \ref{app:prompts_datasets}, Table \ref{tab:full_prompts}). Then, the answer is compared against the gold answer to compute the final answer accuracy.

\textbf{Sensibility and Compliance.}
We classify CoT $\mathbf{\hat{r}}$ with the LLM judge introduced in \S\ref{sec:automated_reasoning_classification}, and compare against $\mathbf{t}$ and $\mathbf{t'}$ to acquire the sensibility and compliance rates respectively (see \S\ref{sec:definiton}).
We use \textsc{GPT-5.1} API with $0.1$ temperature as a judge for models other than \textsc{GPT-5.1}, then use \textsc{Gemini3-flash} to judge \textsc{GPT-5.1}.

\textbf{Confidence Estimation.}
As textual CoT may be unfaithful, it is necessary to also analyze signals beyond textual generation.
We start with measuring the confidence for generating answer $\mathbf{a'}$ to quantify how the reasoning conflict impact the LLM's internal states.
$P($True$)$ \citep{kadavath2022languagemodelsmostlyknow} let an LLM generate a response with the original prompt (the reasoning instructions in our case), then ask a follow-up question in the same conversation: Is the answer \textit{(A) True} or \textit{(B) False}. The confidence score can thus be extracted as the probability of generating the token \textit{A}. The full prompt details are shown in Appendix~\ref{app:prompts_datasets}.

\textbf{Probing for Reasoning.}
To examine whether reasoning instructions are encoded in the model's internal states, we apply linear probing to the residual stream activations of the open LLMs. For each response generated under the reasoning instructions, we collect hidden states from every transformer layer after the feed-forward block, restrict the representation to the question span, and obtain one vector per layer by mean-pooling over the span tokens. On top of these frozen representations, we train independent multi-class linear probes to predict: 1) the instructed reasoning type, 2) the judge-inferred reasoning type, and 3) the reasoning is compliant or not. 
Probe performance is measured layer-wise using accuracy and macro-F1, and we summarize each model family by the strongest probe layer and by the full layer-wise curves.

\begin{figure*}[t]
    \centering
    \includegraphics[width=0.8\textwidth]{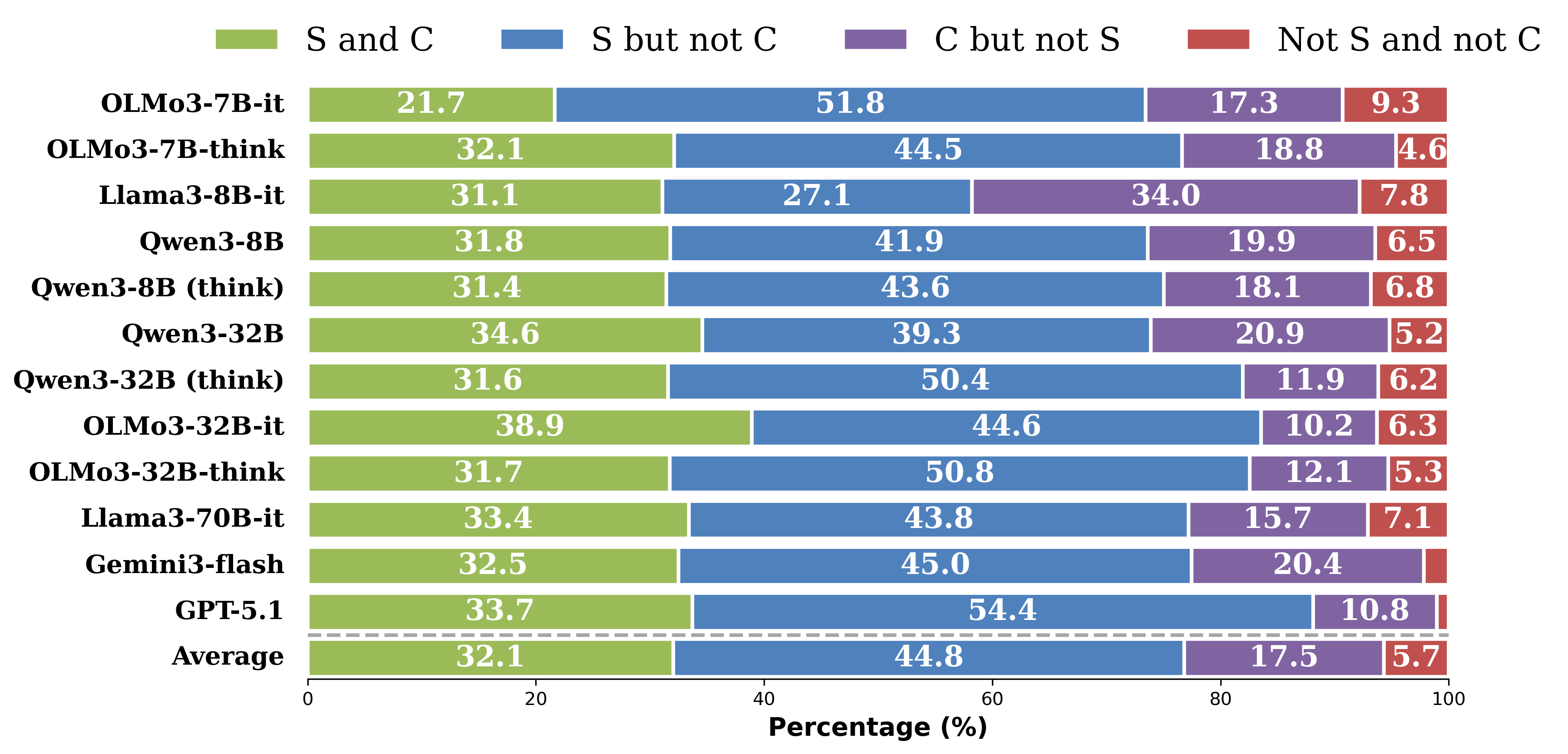}
    \caption{Whether the reasoning is sensible (S) or compliant (C) based on the LLM judge.}
    \label{fig:sensible_compliant}
\vspace{-8pt}
\end{figure*}

\begin{figure}[t]
    \centering
    \includegraphics[width=0.95\columnwidth]{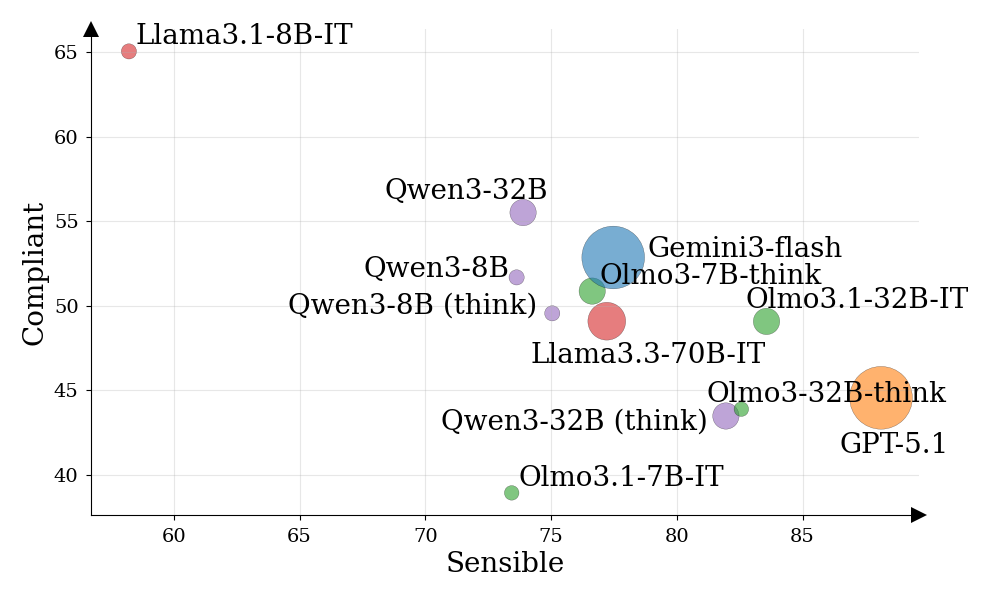}
    \caption{Proportion of sensible vs. compliant CoT across models. Larger circles represent larger LLMs.}
    \label{fig:sensible_compliant_model_family}
    \vspace{-8pt}
\end{figure}

\section{Results}
\subsection{Revealing Difficulty in Controllability}
\label{sec:exp_1}
\textbf{LLMs prioritize logical sensibility over instruction compliance.}
As illustrated in Fig. \ref{fig:sensible_compliant}, our aggregated results across all datasets, all instructed reasoning types, and random seeds reveal a clear hierarchy: LLMs consistently prioritize logical sensibility over instruction compliance.
On average, models opt for the reasoning type that matches the task requirements in 43.5\% of cases ($S \cap \neg C$). In contrast, models exhibit total compliance by following the instructed type at the expense of task appropriateness in only 18.6\% of instances $(\neg S \cap C)$.
We find that stray reasoning, where the model utilizes a framework that is neither sensible nor compliant, is notably rare, occurring in only 5.7\% of the samples. This ``Neither'' category consistently accounts for less than 10\% of total outputs across all tested models. Furthermore, we observe a correlation with model scale: larger models demonstrate a higher degree of ``logical grounding'', reducing the frequency of outputs that fall into the non-sensible, non-compliant quadrant.

\textbf{Reasoning sensibility scales with model sizes, with notable exceptions.}
As shown in Fig. \ref{fig:sensible_compliant_model_family}, reasoning sensibility generally improves with model scale. \textsc{GPT-5.1} exhibits the highest sensibility score ($88.1\%$), whereas \textsc{Llama3.1-8B-IT} is the least sensible ($58.2\%$). However, this scaling trend is not universal. Most notably, the \textsc{Qwen3} with thinking mode off demonstrates nearly identical sensibility rates ($73.5\%$ vs. $73.9\%$), despite \textsc{Qwen3-32B} possessing a four-fold increase in parameters over the 8B variant. Moreover, \textsc{Gemini3-flash} underperforms relative to its parameter class, trailing behind \textsc{Llama3.3-70B-IT} and \textsc{Olmo3.1-32B-IT}. These results suggest that while increased sizes typically fosters more sensible reasoning, architectural choices and specific instruction-tuning recipes play an essential role in determining a model's logical grounding.

\textbf{Instruction compliance is family-dependent and parameter-agnostic.}
Our results (Fig. \ref{fig:sensible_compliant_model_family}) indicate that compliance rates diverge greatly across model families. Model scale serving as an unreliable predictor of instruction compliance. \textsc{Llama3.1-8B-IT} achieves a substantially higher compliance rate ($65.1\%$) than all other evaluated models, while \textsc{Olmo3-7B-IT} exhibits the lowest ($39\%$). Within the \textsc{Qwen3} and \textsc{Olmo3} lineages, we observe standard scaling, where the 32B variants outperform their 8B/7B counterparts in compliance. Conversely, the \textsc{Llama} family demonstrates an inverse trend: \textsc{Llama3.1-8B-IT} is notably more compliant than the much larger \textsc{Llama3.3-70B-IT}. This high variance across families and the non-monotonic relationship with model size suggest that compliance is primarily driven by specific post-training methodologies rather than their scales.

\textbf{Models with build-in thinking are generally more sensible and less compliant.}
Based on Fig. \ref{fig:sensible_compliant} and \ref{fig:sensible_compliant_model_family}, \textsc{Qwen} models have higher sensible rate when turning the thinking mode on.
On the other hand, switching thinking mode on makes the compliant rate lower.
The larger thinking \textsc{Olmo} also has a lower compliant rate.
This indicate a substantial behavior difference between the build-in thinking and the CoT outside of the thinking tags. 

\begin{figure*}[ht]
    \centering
    \includegraphics[width=\textwidth]{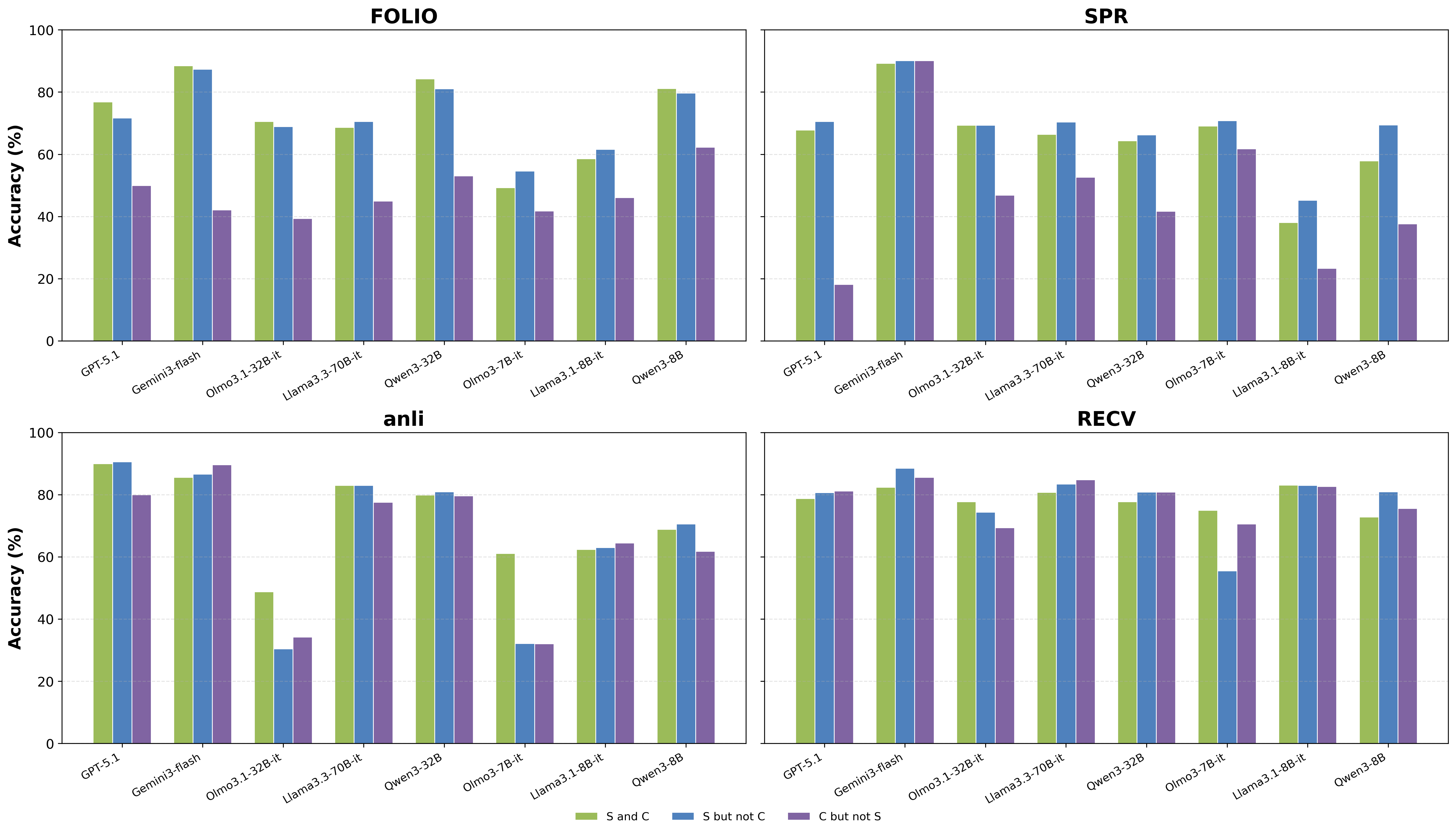}
    \caption{The average accuracies of the final answers with respect to the categories.} 
\label{fig:accuracy_wrt_sensible_compliance_category}
\vspace{-8pt}
\end{figure*}

\textbf{Logical sensibility is the primary determinant of task accuracy.}
As illustrated in Fig. \ref{fig:accuracy_wrt_sensible_compliance_category}, LLMs consistently achieve higher accuracy when their reasoning is sensible, regardless of instruction compliance (i.e., the sensible/compliant and sensible/non-compliant cohorts). Due to its limited frequency, we exclude the fourth category (not sensible/not compliant) from this analysis as it is not sufficient for statistical significance testing (Fig. \ref{fig:sensible_compliant}). This disparity is most pronounced on the FOLIO dataset, where sensible CoT maintains a $20\%$--$40\%$ accuracy lead over the compliant/not sensible category. Conversely, the \textsc{RECV} dataset exhibits marginal gaps, with differences across all categories falling below $20\%$. At the model level, \textsc{Gemini3-flash} displays relatively uniform accuracy across the three cohorts, with the notable exception of its compliant/not sensible performance on FOLIO. This suggests that \textsc{Gemini3-flash} may rely more heavily on internal heuristics, with its generated CoT serving primarily as post-hoc justification rather than functional reasoning steps. In Appendix \ref{sec:more_accuracy_results}, we show the numeric accuracies in Table \ref{tab:sensible_and_compliance_accuracy}, and the accuracies grouped by the instructed reasoning types in Table \ref{tab:results_wrt_instructed_type}.

\subsection{Confidence Reflects Internal Reasoning Conflicts}
\textbf{Instruction compliance is positively correlated with model confidence.}
To look beyond the textual reasoning chains induced by our instructions, we investigate the relationship between reasoning compliance and the internal confidence of the LLMs. We quantify confidence using the $p(\text{True})$ probability of the final answer, aggregated across the six open-weight instruction LLMs over three instructions and three random seeds per dataset. To isolate the impact of compliance from the correlation between accuracy and confidence, we marginalize the effect of correctness by grouping results into correct and incorrect cohorts (Table \ref{tab:confidence_overall}).

Our analysis reveals that $p(\text{True})$ confidence of the compliant responses is significantly higher than the non-compliant ones ($p < 0.05$), regardless of the final answer's correctness. 
However, the effect size (Cohen's $d$) is $0.09$ on the correct responses and $0.12$ on the incorrect responses.
These results indicate that reasoning conflicts do not cause catastrophic drop in model confidence, but they introduce a subtle frictional drag that is mathematically verifiable during the response generation.
The significance in the confidence drop during non-compliant episodes still implies that models maintain an internal awareness of the provided instructions, which imposes an implicit influence on the generation process even when the model ultimately defaults to a different reasoning paradigm.
This motivates us to further conduct mechanistic analysis of the internal activation.

\begin{table}[!t]
\centering
\scriptsize
\resizebox{0.9\columnwidth}{!}{%
\begin{tabular}{lcccc}
\toprule
\textbf{Correctness} & \textbf{Compliant} & \textbf{Not Compliant} & $\Delta$ & $p$ value\\
\midrule
Incorrect & 0.5664 & 0.4831 & 0.0833 & 3e-14\\
Correct & 0.7116 & 0.6937 & 0.0179 & 5e-13\\
\bottomrule
\end{tabular}%
}
\caption{Average $p(\textit{True})$ confidence scores of compliant reasoning and not compliant reasoning grouped by incorrect or correct final answers.}
\label{tab:confidence_overall}
\vspace{-8pt}
\end{table}

\begin{figure*}[ht]
    \centering
    \includegraphics[width=0.95\textwidth]{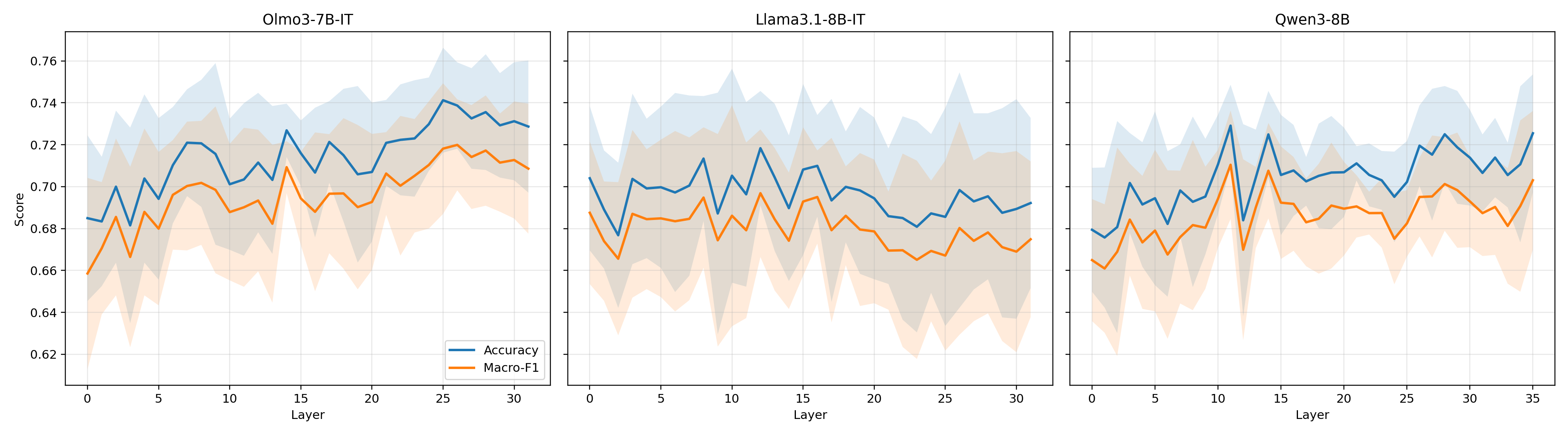}
    \caption{The probing scores across layers for complicant vs. non-compliant binary classification.}% (random baseline is $0.5$).}
    \label{fig:three_probing_graphs}
\vspace{-8pt}
\end{figure*}

\subsection{Mechanistic Analysis}
\label{sec:mechanistic}
Fig. \ref{fig:three_probing_graphs} shows reasoning compliance is moderately decodable from the model's hidden states, with decoding accuracy typically peaking in the middle and late layers for \textsc{Olmo} and \textsc{Qwen}. This suggests that the signal for compliance strengthens as the forward pass progresses.
In our preliminary experiment, we also probe the instructed reasoning type and judge-inferred reasoning type (Appendix \ref{app:probing}). Compared to them, compliance is a weaker and less linearly accessible property of the internal state. These results show that although the LLMs robustly encode the instructed reasoning types, whether they eventually comply with it is a weaker and more entangled property of the internal state.

\begin{figure}[t]
    \centering
    \includegraphics[width=\columnwidth]{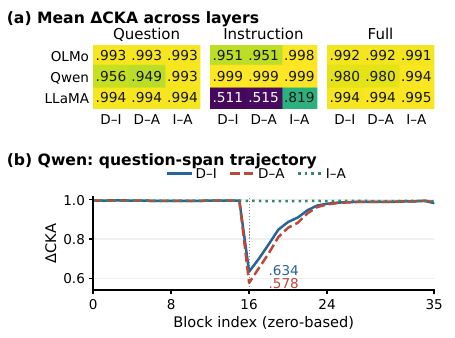}
    \caption{Mean $\Delta$CKA across spans (a) and Qwen's question trajectory (b).
    D/I/A: deduction/induction/abduction; higher values indicate greater alignment
    beyond permutation baseline.}
    \label{fig:cka_compact}
\end{figure}

\paragraph{Instruction-dependent geometry.}
We complement probing with CKA~\citep{kornblith2019similarity} between
same-question representations under different reasoning instructions,
measured over instruction, question, and full-prompt spans
(Fig.~\ref{fig:cka_compact}; Appendix~\ref{app:cka}).
LLaMA shows the largest instruction-span differences
($\Delta$CKA $0.511/0.515$ for deductive--inductive/abductive), including
an inductive--abductive difference ($0.819$); OLMo mainly distinguishes
deduction ($0.951$), whereas Qwen retains high alignment ($0.999$).
Question and full-prompt geometry remain highly similar in LLaMA and OLMo.
Qwen instead exhibits a transient question-span dip for deduction
contrasts at block~16 ($0.634/0.578$), with a smaller full-prompt effect
and substantial recovery in later blocks.

This distinguishes instruction decodability from changes in overall
geometry: high question-span CKA coexists with near-perfect instruction
decoding, including in non-compliant responses. Qwen's divergence is
consistent with its stronger judge-label probe signal, while
LLaMA's high compliance despite stable question geometry cautions
against treating geometric divergence as a compliance score.
Together with steering's task-dependent effects,
these results distinguish instruction representation, geometric change,
and behavioural controllability; neither CKA nor decodability alone
establishes causal use of the instructions or CoT faithfulness.

\begin{figure*}[!t]
     \centering
     \begin{subfigure}[b]{0.32\textwidth}
         \centering\includegraphics[width=\textwidth]{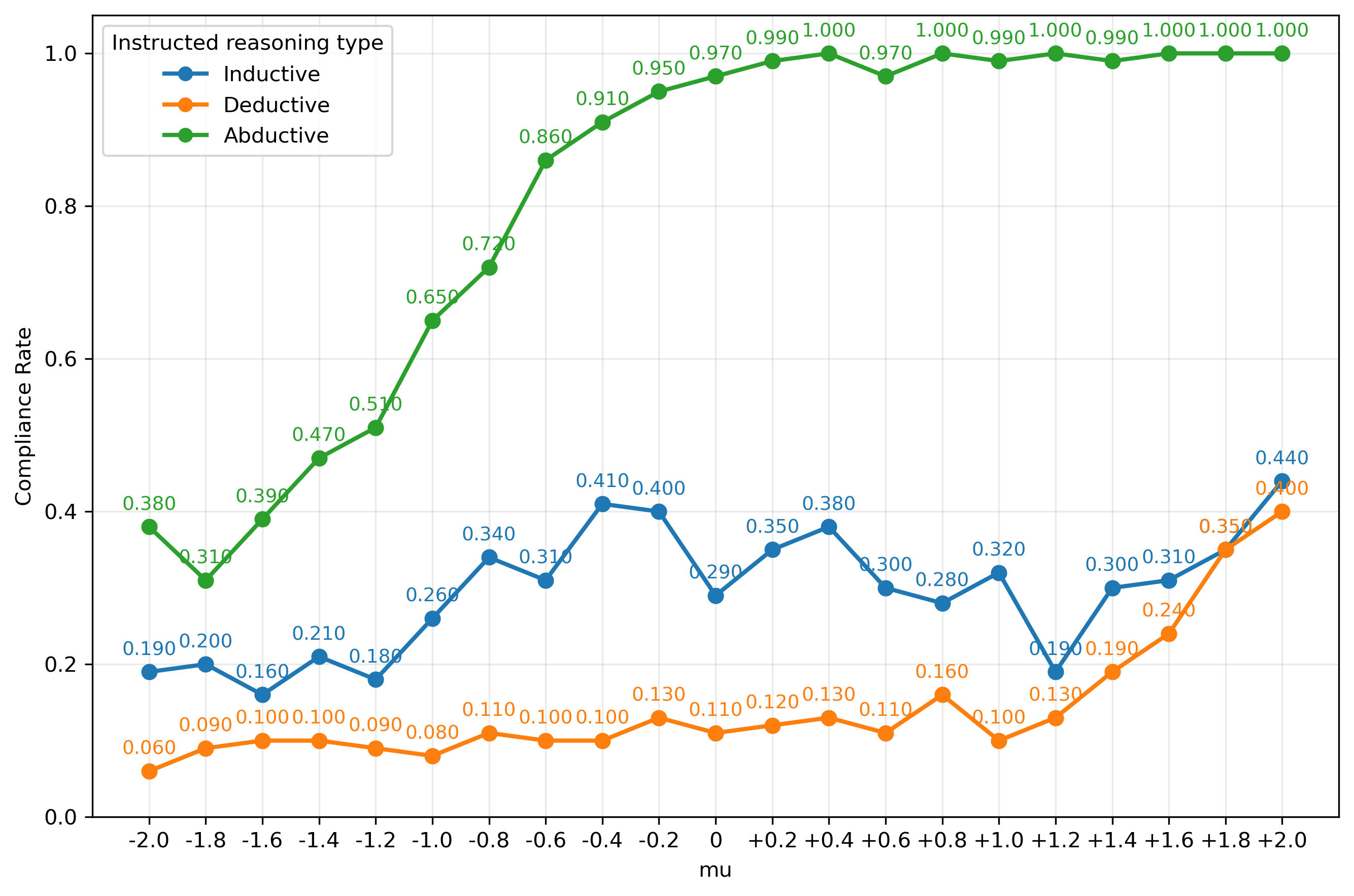}
         \caption{Compliance Rate}
         \label{fig:compliance_steer_olmo}
     \end{subfigure}
     %\hfill
     \begin{subfigure}[b]{0.32\textwidth}
         \centering
         \includegraphics[width=\textwidth]{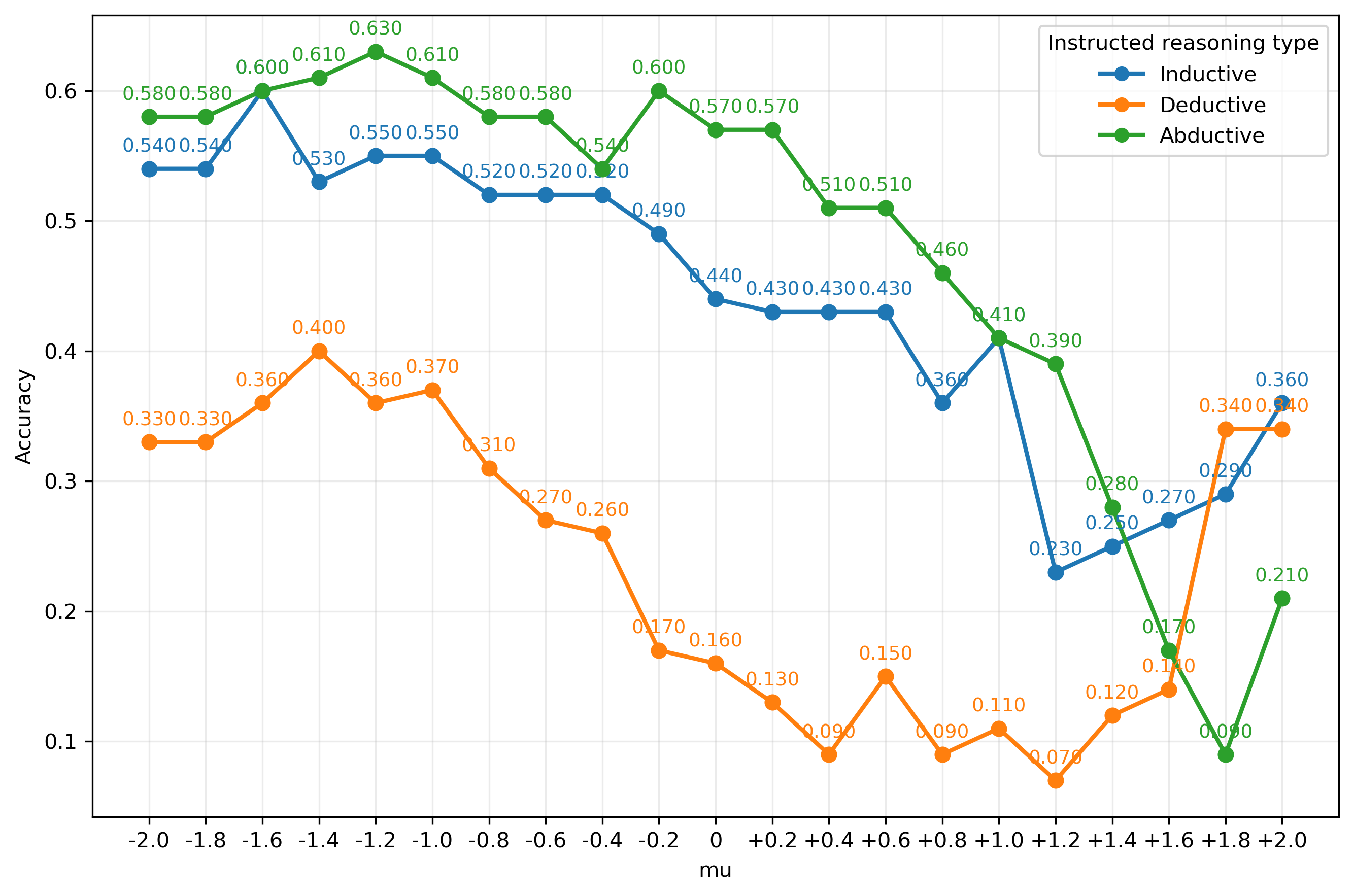}
         \caption{Final Answer Accuracy}
         \label{fig:accuracy_steer_olmo}
     \end{subfigure}
     %\hfill
     \begin{subfigure}[b]{0.32\textwidth}
         \centering
         \includegraphics[width=\textwidth]{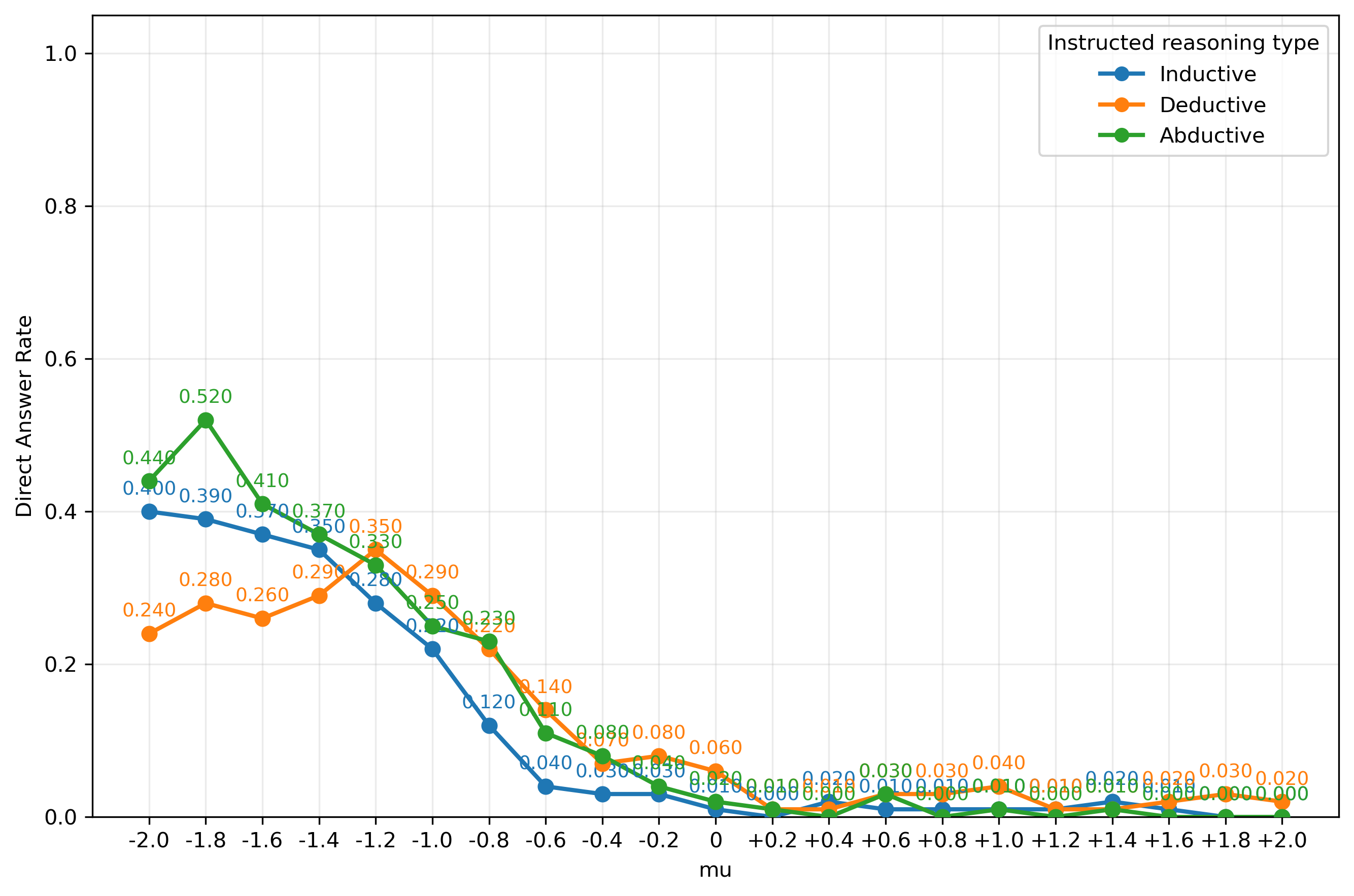}
         \caption{Direct Answer Ratio.}
         \label{fig:direct_steer_olmo}
     \end{subfigure}
        \caption{The impact of multiplier $\mu$ on $\alpha$-NLI when steering the layer 14-17 of \textsc{Olmo3-7B-IT}.}
        \label{fig:steering_olmo_anli}
\vspace{-8pt}
\end{figure*}

\begin{figure*}[ht]
     \centering
     \begin{subfigure}[b]{0.32\textwidth}
         \centering
         \includegraphics[width=\textwidth]{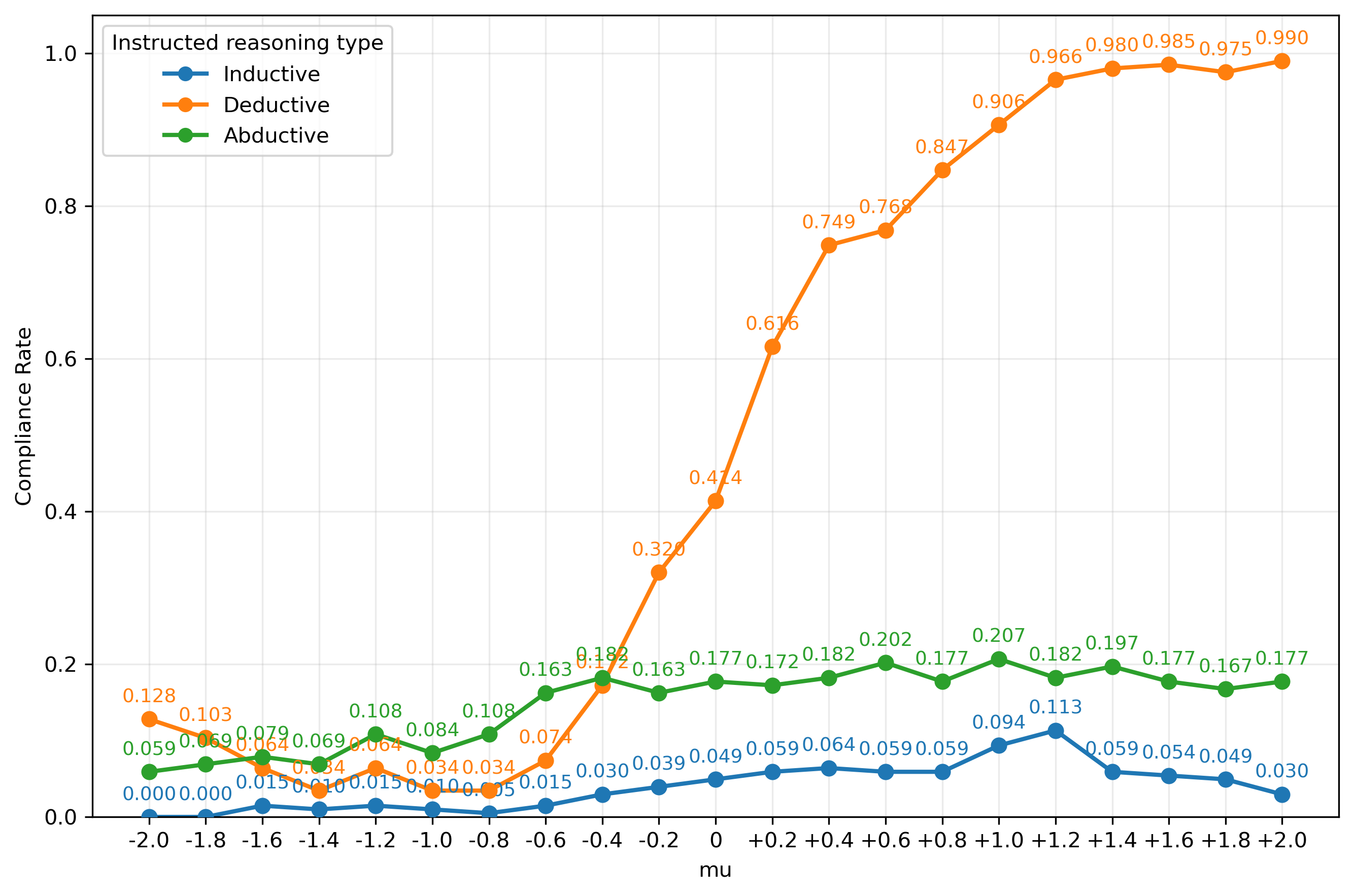}
         \caption{Compliance Rate}
         \label{fig:compliance_steer_olmo_folio}
     \end{subfigure}
     \hfill
     \begin{subfigure}[b]{0.32\textwidth}
         \centering
         \includegraphics[width=\textwidth]{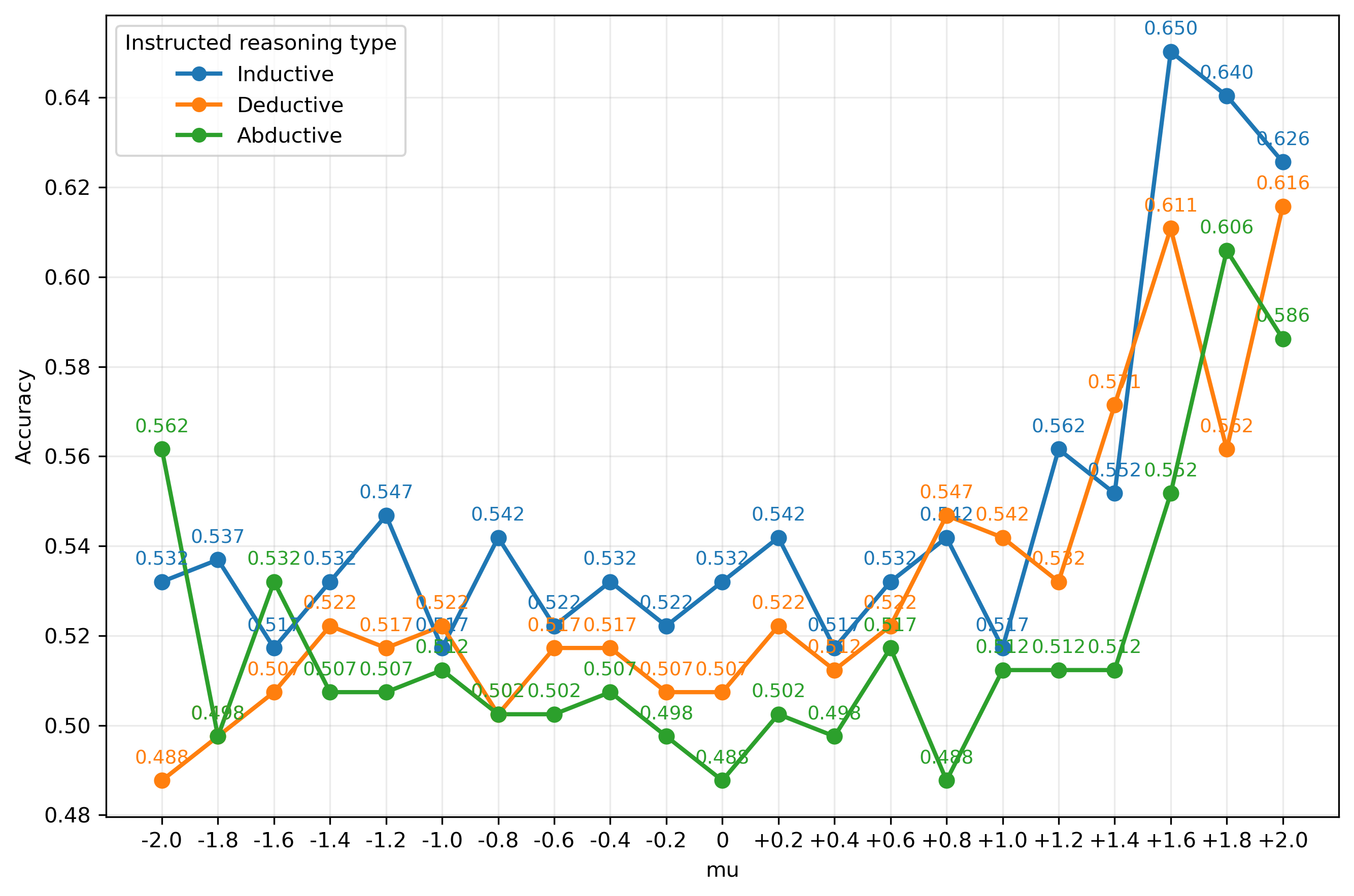}
         \caption{Final Answer Accuracy}
         \label{fig:accuracy_steer_olmo_folio}
     \end{subfigure}
     \hfill
     \begin{subfigure}[b]{0.32\textwidth}
         \centering
         \includegraphics[width=\textwidth]{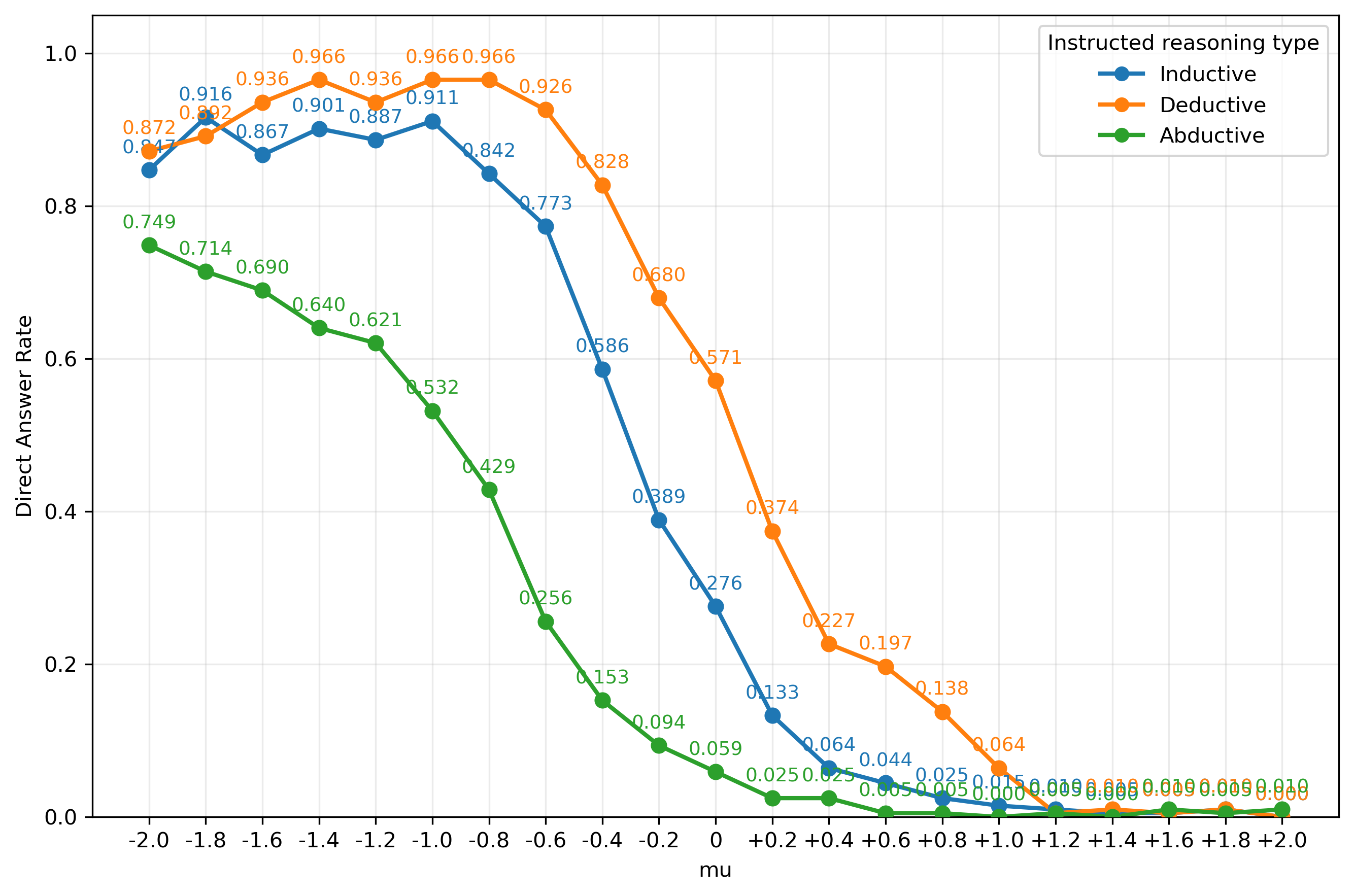}
         \caption{Direct Answer Ratio}
         \label{fig:direct_steer_olmo_folio}
     \end{subfigure}
        \caption{The impact of multiplier $\mu$ on \textsc{FOLIO} when steering the layer 14-17 of \textsc{Olmo3-7B-IT}.}
        \label{fig:steering_olmo_folio}
\end{figure*}

%Based on the probing results (Fig. \ref{fig:three_probing_graphs} and Appendix \ref{app:probing}), we can identify the middle layers as the primary site for encoding reasoning paradigms, we applied CAA to layers 14--17 of \textsc{Olmo3-7B-IT}. 
\subsection{Enhancing Compliance via Steering}
\label{sec:steering}
Guided by the probing results (Fig.~\ref{fig:three_probing_graphs}; Appendix~\ref{app:probing}), we applied CAA to layers 14--17 of \textsc{Olmo3-7B-IT} to test whether observed compliance could be shifted.This model was specifically chosen for intervention due to its baseline status as the least compliant model in our study (Fig. \ref{fig:sensible_compliant_model_family}). As shown in Fig. \ref{fig:compliance_steer_olmo}, applying steering vectors with a positive multiplier ($\mu$) greatly increases the compliance rate, with the effect scaling alongside $\mu$. This is particularly evident when forcing \textit{inductive} and \textit{deductive} instructions on $\alpha NLI$ questions, where steering successfully overrides the model’s natural abductive inclination. However, the impact on task accuracy (Fig. \ref{fig:accuracy_steer_olmo}) reveals a complex trade-off. For abductive instructions, accuracy drops as $\mu$ increases; qualitative analysis suggests this is due to a ``task-neglect'' side effect, where the model becomes so focused on the steered reasoning type that it fails to follow formatting constraints (e.g., enclosing the final answer in the required tags). Interestingly, for \textit{inductive} and \textit{deductive} instructions, accuracy follows a non-monotonic trend: initially declining before reaching an upward peak as compliance is enforced. Finally, we observe that steering with a negative $\mu$ consistently improves accuracy across all instruction types.
Such ``anti-steering'' also makes LLMs resort to direct answer (Fig. \ref{fig:direct_steer_olmo}), potentially reducing the cognitive friction caused by conflicting reasoning.
The steering results on \textsc{FOLIO} (Fig. \ref{fig:steering_olmo_folio}) also show a generally similar trend, where the compliance rate of deductive instruction is substantially increased.
Moreover, the accuracies on all three instructed reasoning type are greatly improved with larger $\mu$.
More steering results on other datasets and models are in Appendix \ref{app:more_steering}.

\subsection{Case Study}
\label{sec: case_study}

Example 1 shows the output of \textsc{Llama3.3-70B-it} when explicitly instructed to use abductive reasoning. Notably, the model exhibits a form of lexical mirroring: it incorporates key abductive terminology from the prompt (e.g., ``most probable'' and ``simplest hypothesis'') to signal compliance. However, the internal logic later pivot to deductive logical entailment to reach the final answer.
This suggests that while the model knows the requested reasoning is inappropriate for the task, its true reasoning process is masked behind a compliant textual facade. Our LLM judge correctly identifies the reasoning type as \textit{deductive}, despite the model's self-proclamation of abduction. 

\begin{tcolorbox}[
    breakable,
    colback=green!5,
    colframe=green!60!black,
    title=Example 1
]
    \small
    To determine the validity of the conclusion ``Montana is home to the city of Missoula'' based on the given premises, let's analyze each premise step by step and apply abductive reasoning to find the most probable and simplest hypothesis that explains all observations. [...] Given the direct statement in premise 2 that ``The state of Montana includes the cities of Butte, Helena, and Missoula'', we have explicit information supporting the conclusion that Montana is home to the city of Missoula. The other premises either provide additional context that does not contradict this conclusion or introduce information about other locations that do not directly impact the conclusion about Missoula. Therefore, based on the premises provided, the conclusion that ``Montana is home to the city of Missoula'' is supported directly by premise 2 and is not contradicted by any other premise.
\end{tcolorbox}

\section{Conclusion}
We characterise reasoning conflicts through behavioural evaluation, probing, representation similarity, and activation steering. Instructed reasoning types remain decodable even in non-compliant responses, while CKA reveals geometric differences across models and spans. Steering increases observed compliance by up to $29\%$, with task-dependent effects on accuracy and output quality. Together, these findings distinguish instruction encoding from behavioural compliance and establish a degree of activation-level controllability. Future work should test their causal role and evaluate questions requiring multiple reasoning types.

\section*{Limitations}
While CAA proves capable of steering model compliance, it does so at a visible cost to generation quality. Our intervention experiments are primarily validated on localized open-weights architectures where strong activation steering frequently induced formatting degradation or task neglect. Future research is required to evaluate whether scaling laws or dynamic, layer-specific steering vectors can isolate reasoning compliance without destabilizing the model's core textual coherence.

\section*{Acknowledgments}
XT, MA, YZ, ML and NA are supported by the EPSRC [grant number EP/Y009800/1], through funding from Responsible AI UK (KP0016) as a Keystone project.
The authors acknowledge the use of resources provided by the Isambard-AI National AI Research Resource (AIRR) \cite{mcintoshsmith2024isambardaileadershipclasssupercomputer}. Isambard-AI is operated by the University of Bristol and is funded by the UK Government’s Department for Science, Innovation and Technology (DSIT) via UK Research and Innovation; and the Science and Technology Facilities Council [ST/AIRR/I-A-I/1023].
The authors also acknowledge IT Services at the University of Sheffield and the University of Oxford Advanced Research Computing for the provision of HPC services.
The authors also thank Atsuki Yamaguchi for his valuable help and feedback.

% Bibliography entries for the entire Anthology, followed by custom entries
%\bibliography{anthology,custom}
% Custom bibliography entries only
\bibliography{custom}

\clearpage

\appendix

\section{Implementation Details}
We use vLLM \citep{kwon2023efficient} on NVIDIA GH200 (96G), A100 (80G), and AMD MI300X (192G) for generating the responses with open LLMs.
The inference temperature is set to $0.5$.
We sample responses from LLMs with $3$ different random seeds.
For \textsc{Qwen3-8B} and \textsc{Qwen3-32B}, the thinking mode is disable.
We use \textit{EasyEdit 2.0} for model steering \citep{wang-etal-2024-easyedit,zhang2024comprehensive}.
For probing, we use a default logistic regression classifier with $C=1.0$ and \textsc{L2} penalty.

\section{Dataset Details}
For Sequence Pattern Recognition (SPR), we use the code of \citet{hu2025ahasystematicmetaabilitiesalignment} to generate a training set of $4,300$ samples and a test set of $100$ samples.
We randomly split RECV \cite{dougrez-lewis-etal-2025-assessing} by stratifing the labels in to train, dev, and test sets.
Each subset (i.e., climate fever, VitC, and PhemePlus) has the exact same number of samples.
This results in $99$ test samples and $1,302$ train samples.
We randomly sample $100$ questions from the test set of $\boldsymbol{\alpha}$NLI to have a balanced number of test samples as the other datasets.
We use the training set of FOLIO, $\boldsymbol{\alpha}$NLI, and SPR for generating the steering vectors.
For all the evaluation, we use the test split of $\boldsymbol{\alpha}$NLI, SPR, and RECV, and the validation split of FOLIO as it does not have a test split available.

\begin{table*}[ht] 
\scriptsize % Slighting smaller font to save space
\centering
\begin{tabularx}{\textwidth}{l X}
\toprule
\textbf{Type} & \textbf{Prompt} \\
\midrule
\textbf{Deductive} & You are a logician tasked with performing \textbf{deductive} reasoning. You are given a general rule and specific observations. Your task is to apply the general rule to the observations to derive a logically certain conclusion. Provide a detailed reasoning process leading to your conclusion. [Question, Context, and Options...] Please enclose the answer in $<$answer$>$$<$/answer$>$.\\
\midrule
\textbf{Inductive} & You are a logician tasked with performing \textbf{inductive} reasoning. You are given a set of observations. Your task is to infer the most probable general rule that explains all observations. Then, use the inferred rule to make predictions about new observations. Provide a detailed reasoning process leading to your conclusion. [Question, Context, and Options...] Please enclose the answer in $<$answer$>$$<$/answer$>$.\\
\midrule
\textbf{Abductive} & You are a logician tasked with performing \textbf{abductive} reasoning. You are given a general rule and an observation. Your task is to generate the most probable and simplest hypothesis that, if true, would logically explain all the observations provided. Provide a detailed reasoning process leading to your conclusion. [Question, Context, and Options...] Please enclose the answer in $<$answer$>$$<$/answer$>$. \\
\bottomrule
\end{tabularx}
\caption{Prompts for controlling LLMs' reasoning.}
\label{tab:full_prompts}
\end{table*}

\section{Prompt Details}
\label{app:prompts_datasets}
This section presents the complete prompts used in our evaluation pipeline.
Table \ref{tab:full_prompts} show the full prompts for inducing reasoning conflicts, including the special tags we ask the models to add for the answer matching.

Below is the prompt for $p(\textit{True})$ confidence estimation:
\begin{tcolorbox}[
    breakable,
    colback=yellow!5,
    colframe=yellow!60!black,
    title=$p(\textit{True})$ Prompt
]
    User: [Question + Reasoning Instruction]\\
    LLM: [CoT + Final Answer]\\
    User:\\
    Is this answer:\\
    (A) True\\
    (B) False\\
    The answer is: (
\end{tcolorbox}

\begin{table}[ht]
\centering
\resizebox{0.75\columnwidth}{!}{%
\begin{tabular}{lcc}
\toprule
\textbf{Model} & \textbf{Accuracy} & \textbf{Cohen's $\kappa$} \\
\midrule
GPT-5.1 & .89 & .84 \\
Gemini-3-flash & .88 & .83 \\
\bottomrule
\end{tabular}%
}
\caption{Agreement between the LLM judges and the human annotators.}
\label{tab:judge_verification}
\end{table}

\begin{table}[!t]
    \centering
    \resizebox{0.5\columnwidth}{!}{
    \begin{tabular}{lc}
    \toprule
        $\mu$ & Compliance Rate \\
        \midrule
        $-1.0$ & $24.2$ \\
        $0$ & $81.8$    \\
        $1.0$ & $87.9$   \\
        \bottomrule
    \end{tabular}
    }
    \caption{Human evaluation of the compliance rate with respect to the steering vector multiplier $\mu$.}
    \label{tab:human_evaluation_steering_mu}
\end{table}

\begin{figure}[ht]
    \centering
    \includegraphics[width=0.5\columnwidth]{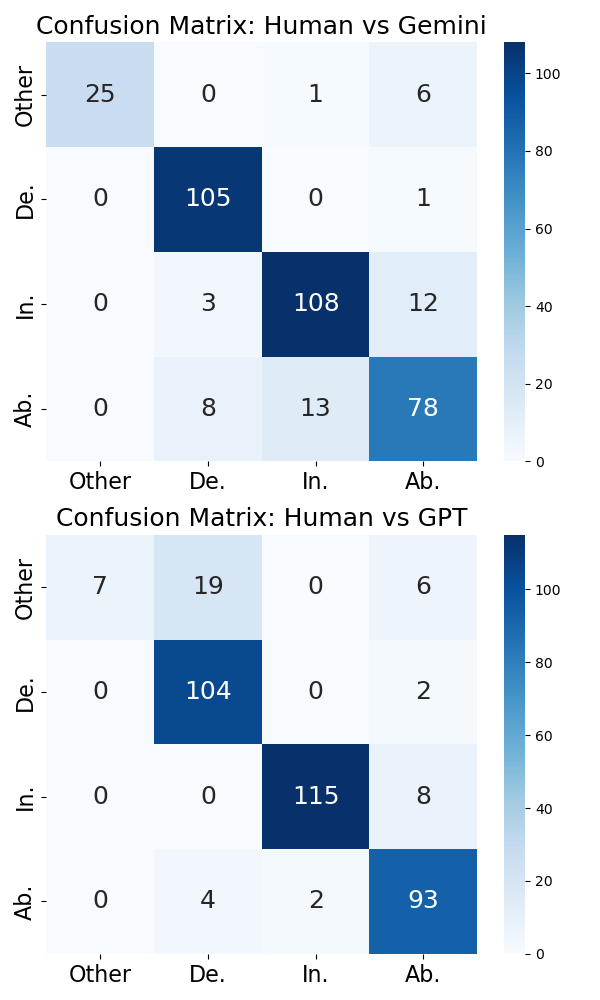}
    \caption{The confusion matrix between the aggregated human labels and the LLMs' labels. The rows represents human predictions.}
    \label{fig:confusion_matrix_IAA}
\end{figure}

% Insert after original line 693, before the following accuracy-table float.
\section{Representation Similarity across Reasoning Instructions}
\label{app:cka}

\paragraph{Setup.}
We compare the same questions under deductive, inductive, and abductive
instructions, aligning inputs by question content. At each transformer-block
output, after the feed-forward sublayer, we mean-pool over instruction,
question, or full-prompt tokens. PCA retains 256 components separately
for each condition and block. We compute linear centred kernel alignment
on these representations and subtract the mean of 20 random permutations
of question correspondence:
\begin{equation}
\Delta\mathrm{CKA}_{\ell}
=\mathrm{CKA}_{\ell}^{\mathrm{paired}}
-\frac{1}{20}\sum_{b=1}^{20}\mathrm{CKA}_{\ell,b}^{\mathrm{perm}}.
\end{equation}
Higher values indicate greater alignment beyond this baseline, rather than
more instruction information. Block indices are zero-based: 0--31 for
OLMo/LLaMA and 0--35 for Qwen. The question/full-prompt analyses retain
510 aligned questions for OLMo, 890 for Qwen, and 889 for LLaMA;
cross-model comparisons are therefore descriptive.

\begin{table}[!ht]
\centering
\small
\setlength{\tabcolsep}{3pt}
\begin{tabular}{llrrr}
\toprule
Model & Span & D--I & D--A & I--A \\
\midrule
OLMo & Question & .9933 & .9932 & .9932 \\
     & Instruction & .9507 & .9507 & .9979 \\
     & Full & .9916 & .9916 & .9914 \\
\midrule
Qwen & Question & .9561 & .9487 & .9932 \\
     & Instruction & .9990 & .9990 & .9990 \\
     & Full & .9803 & .9798 & .9944 \\
\midrule
LLaMA & Question & .9936 & .9937 & .9939 \\
      & Instruction & .5109 & .5149 & .8188 \\
      & Full & .9940 & .9944 & .9947 \\
\bottomrule
\end{tabular}
\caption{Layer-averaged $\Delta$CKA for OLMo3-7B-IT,
Qwen3-8B, and LLaMA3.1-8B-IT. D/I/A denote
deductive/inductive/abductive instructions.}
\label{tab:cka_all_spans}
\end{table}

\paragraph{Span and layer differences.}
Table~\ref{tab:cka_all_spans} reports all comparisons. Deduction is the
most consistent axis of geometric difference, but LLaMA also shows
lower instruction-span alignment between induction and abduction.
These instruction-span contrasts concern the geometry of the instruction
positions and can reflect wording; they do not quantify semantic
understanding. Full-prompt pooling can attenuate local differences,
so its high similarity does not exclude changes at individual positions.
In Qwen's question span, raw CKA falls from $0.9995/0.9997$ at block~15
to $0.6390/0.5826$ at block~16 for deductive--inductive/abductive
comparisons. The dip is therefore present before baseline subtraction.
By block~34, $\Delta$CKA recovers to $0.9933/0.9937$; the effect is
transient rather than a persistent separation throughout later blocks.

\paragraph{Relation to probing and steering.}
Centred CKA can remain high when instruction identity is readily
decodable: a condition-specific offset is removed by centring, and
changes in a limited subspace need not dominate the global geometry.
Accordingly, high question-span CKA does not mean instruction information
failed to propagate, while lower CKA does not identify a particular
reasoning algorithm. OLMo's high CKA alongside steering-induced compliance
changes further separates baseline geometric similarity from
manipulability. These analyses complement the behavioural and probing
results, but do not establish a causal link between a particular geometric
difference and the reasoning expressed in the output.

\begin{figure}[t]
    \centering
    \includegraphics[width=\columnwidth]{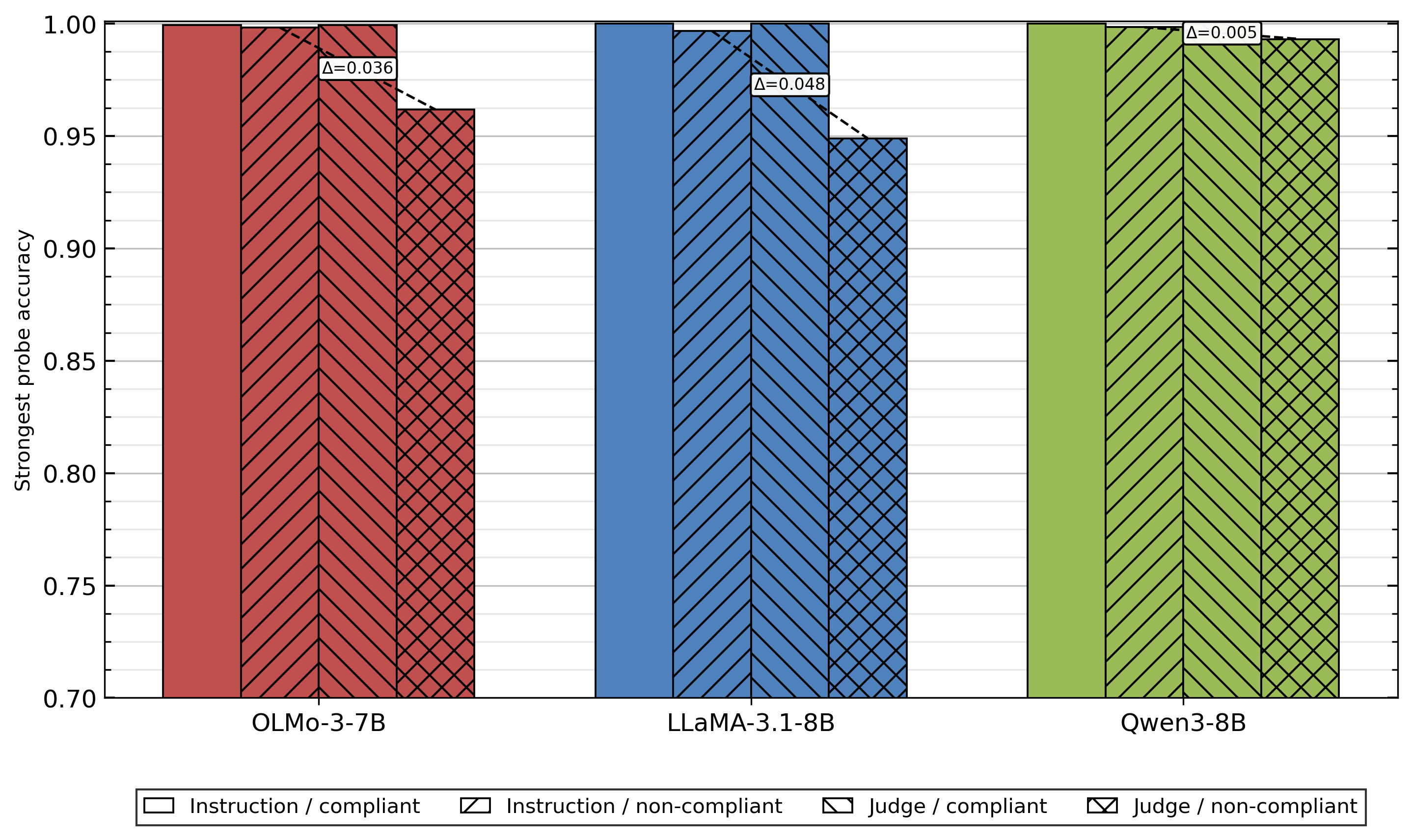}
    \caption{This figure shows the strongest probe accuracy for instructed reasoning type and judge-inferred reasoning type, split by compliant and non-compliant cases, across model families. A larger gap between the non-compliant instructed-reasoning bar and the non-compliant judge-reasoning bar indicates that the model represented the instruction clearly, but did not fully realize it in its final reasoning behavior.}
    \label{fig:probe_family_overlay}
\end{figure}

\section{Probing Results}
 \label{app:probing}
\begin{figure*}[t]
    \centering
    \includegraphics[width=\textwidth]{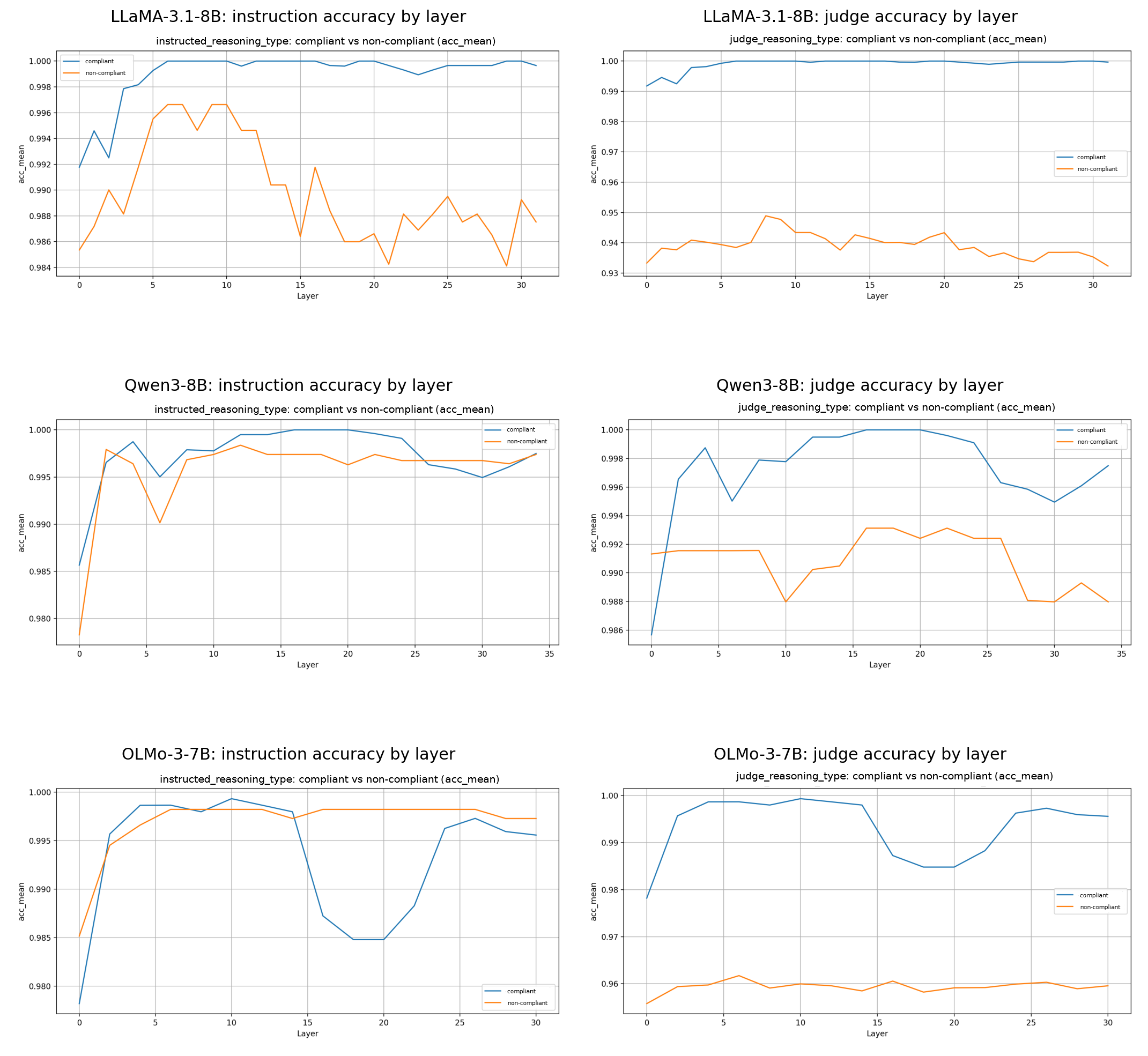}
    \caption{The layer-wise accuracy for each family of model and their instructed and judge-inferred reasoning types.}
    \label{fig:probe_wrt_layers}
\end{figure*}

\textbf{Probing Instruction Encoding and Reasoning Compliance.}
Fig.~\ref{fig:probe_family_overlay} shows our probe analysis across the three open model families. A consistent pattern emerges from the results, the instructed reasoning type is almost perfectly decodable from the residual stream in all families, including on non-compliant cases. This suggests that failures of compliance are not primarily caused by a failure to represent the instruction. Instead, the models appear to encode the prompted reasoning paradigm early and robustly, but may still follow a competing internal trajectory when the instruction conflicts with the question's logical structure. This pattern is strongest in LLaMA, where the gap between instruction decodability and judge-inferred reasoning is largest; \textsc{Olmo} shows the same trend with a smaller gap; and Qwen is the most aligned overall, with both high instruction decodability and the strongest judge-side signal. Taken together, these results support our main finding that the reasoning-instruction conflict is visible at the level of model representations. Instruction identity remains decodable even when the response is non-compliant.
Lower judge-label decodability in non-compliant cases indicates a less linearly accessible signal; it does not independently establish CoT faithfulness.
Fig. \ref{fig:probe_wrt_layers} shows the full layer-wise accuracy results.

\section{LLM Judge Agreement}
\label{app:judge_verification}
To verify the reliability of the automated judge, we prepare a validation dataset from a balanced subset of 360 responses, which are sampled across three model families and questions representing the three fundamental reasoning types.
The annotators are asked to select the dominant reasoning type that directly contribute to the final answer.
Two humans annotate the full validation dataset, then a third annotator reconcile their disagreement to form the gold standard. Both \textsc{Gemini-3-flash} and \textsc{GPT-5.1} demonstrate high alignment with human, achieving Cohen’s kappa ($\kappa$) scores of $0.83$ and $0.84$, respectively (Table \ref{tab:judge_verification}).
To further validate the LLM judge on steered model's outputs, we conduct additional human evaluation on a subset of \textit{OLMo3-7B-IT}'s results from $\alpha$NLI (Fig. \ref{tab:human_evaluation_steering_mu}). The human evaluation results show that the compliance rate rises as the multiplier $\mu$ increases.
Fig. \ref{fig:confusion_matrix_IAA} show the confusion matrix between the aggregated human labels and the LLMs' labels. \textsc{Gemini-3-flash} mostly fails on the abduction type, and often confuses it with the induction type. For \textsc{GPT-5.1}, the confusion between abduction and induction is less common, but it misclassified some deduction type into the ``Other''.
Alternatively, we also test prompting LLM judges with options of the reasoning types and their definitions, but that results in a lower agreement with the humans.
Table \ref{tab:judge_verification} show the accuracy and $\kappa$ of the LLM judges and human annotators.

\begin{table}[!t]
\begin{center}
\resizebox{\columnwidth}{!}{%
\begin{tabular}{clccc}
\toprule

\textbf{Data} & \textbf{LLM} & S and C & S but not C & C but not S \\

\midrule

\multirow{5}{*}{\rotatebox{90}{\footnotesize \textsc{FOLIO}}}
& \textsc{GPT-5.1} & \textbf{76.81} & 71.67 & n.a.  \\ 
& \textsc{Gemini3-flash} & \textbf{88.50} & 87.37 & 42.11 \\ 
\cmidrule{2-5}
& \textsc{Olmo3.1-32B-it}    & \textbf{70.58} & 68.95 & 39.40 \\ 
& \textsc{Llama3.3-70B-it} & 68.67 & \textbf{70.52} & 45.00  \\ 
& \textsc{Qwen3-32B} &       \textbf{84.27} & 81.09 & 53.06 \\ 
\cmidrule{2-5}
& \textsc{Olmo3-7B-it}    & 49.26 & \textbf{54.62} & 41.82 \\ 
& \textsc{Llama3.1-8B-it} & 58.60 & \textbf{61.58} & 46.12 \\ 
& \textsc{Qwen3-8B} &       \textbf{81.14} & 79.66 & 62.32 \\

\midrule

\multirow{5}{*}{\rotatebox{90}{\footnotesize \textsc{SPR}}}
& \textsc{GPT-5.1} & 67.79 & \textbf{70.56} & 50.00\\ 
& \textsc{Gemini3-flash} & 89.23 & 90.09 & \textbf{90.16}\\ 
\cmidrule{2-5}
& \textsc{Olmo3.1-32B-it}    & \textbf{69.32} & 69.32 & 46.88  \\ 
& \textsc{Llama3.3-70B-it} & 66.44 & \textbf{70.42} & 52.63  \\ 
& \textsc{Qwen3-32B} &       64.36 & \textbf{66.28} & 41.72 \\ 
\cmidrule{2-5}
& \textsc{Olmo3-7B-it}    & 69.13 & \textbf{70.84} & 61.81 \\ 
& \textsc{Llama3.1-8B-it} & 38.11 & \textbf{45.24} & 23.35 \\ 
& \textsc{Qwen3-8B} &       57.88 & \textbf{69.41} & 37.65 \\ 

\midrule

\multirow{5}{*}{\rotatebox{90}{\footnotesize $\alpha NLI$}}
& \textsc{GPT-5.1} & 90.00 & \textbf{90.62} & 80.00 \\ 
& \textsc{Gemini3-flash} & 85.57 & 86.67 & \textbf{89.68} \\ 
\cmidrule{2-5}
& \textsc{Olmo3.1-32B-it}    & \textbf{48.82} & 30.48 & 34.21 \\ 
& \textsc{Llama3.3-70B-it} & 83.00 & \textbf{83.01} & 77.57 \\ 
& \textsc{Qwen3-32B} &       79.93 & \textbf{80.95} & 79.61\\ 
\cmidrule{2-5}
& \textsc{Olmo3-7B-it}    & \textbf{61.11} & 32.14 & 32.12 \\ 
& \textsc{Llama3.1-8B-it} & 62.42 & 63.03 & \textbf{64.52} \\ 
& \textsc{Qwen3-8B} &       68.90 & \textbf{70.61} & 61.85 \\ 

\midrule

\multirow{5}{*}{\rotatebox{90}{\footnotesize \textsc{RECV}}}
& \textsc{GPT-5.1} & 78.75 & 80.66 & \textbf{81.18} \\ 
& \textsc{Gemini3-flash} & 82.42 & \textbf{88.54} & 85.63 \\ 
\cmidrule{2-5}
& \textsc{Olmo3.1-32B-it}    & \textbf{77.78} & 74.42 & 69.39  \\ 
& \textsc{Llama3.3-70B-it} & 80.79 & 83.41 & \textbf{84.83} \\ 
& \textsc{Qwen3-32B} &       77.78 & 80.84 & \textbf{80.87} \\ 
\cmidrule{2-5}
& \textsc{Olmo3-7B-it}    & \textbf{75.00} & 55.56 & 70.59 \\ 
& \textsc{Llama3.1-8B-it} & \textbf{83.06} & 83.04 & 82.63  \\ 
& \textsc{Qwen3-8B} &       72.84 & \textbf{80.94} & 75.58 \\ 

\bottomrule
\end{tabular}%
}
\small\caption{The average accuracies of the sensible and compliant reasoning.} 
\label{tab:sensible_and_compliance_accuracy}
\end{center}
\end{table}

\begin{table}[ht]
\begin{center}
\resizebox{0.95\columnwidth}{!}{%
\begin{tabular}{clcccc}
\toprule

\textbf{Data} & \textbf{LLM} & \textsc{direct} & \textsc{deductive} & \textsc{inductive} & \textsc{abductive} \\

\midrule

\multirow{8}{*}{\rotatebox{90}{\footnotesize \textsc{FOLIO}}}
& \textsc{GPT-5.1} & \heatcell{g3} \underline{70.61} \textsubscript{0.8} & \heatcell{g4} \textbf{76.68} \textsubscript{0.6} & \heatcell{g3} 70.44 \textsubscript{1.1} & \heatcell{g3} 71.10 \textsubscript{1.4} \\ 
& \textsc{Gemini3-flash} & \heatcell{g7} \underline{88.01} \textsubscript{0.7} & \heatcell{g7} \textbf{88.51} \textsubscript{1.3} & \heatcell{g4} 79.47 \textsubscript{1.2} & \heatcell{g5} 83.25 \textsubscript{1.3} \\ 
\cmidrule{2-6}
& \textsc{Olmo3.1-32B-it} & \heatcell{g1} 63.05 \textsubscript{0.0} & \heatcell{g4} \textbf{76.85} \textsubscript{1.1} & \heatcell{g1} 62.89 \textsubscript{0.2} & \heatcell{g2} \underline{65.35} \textsubscript{0.8} \\ 
& \textsc{Llama3.3-70B-it} & 59.28 \textsubscript{0.2} & \heatcell{g3} \underline{68.97} \textsubscript{1.1} & \heatcell{g2} 67.82 \textsubscript{1.3} & \heatcell{g3} \textbf{69.13} \textsubscript{0.6} \\ 
& \textsc{Qwen3-32B} & \heatcell{g4} 71.43 \textsubscript{0.4} & \heatcell{g6} \textbf{85.55} \textsubscript{0.5} & \heatcell{g6} \underline{83.42} \textsubscript{1.9} & \heatcell{g5} 81.94 \textsubscript{1.2} \\ 
\cmidrule{2-6}
& \textsc{Olmo3-7B-it} & \textbf{54.35} \textsubscript{1.8} &  \underline{53.69} \textsubscript{0.4} & 52.96 \textsubscript{0.7} &  50.08 \textsubscript{1.6} \\ 
& \textsc{Llama3.1-8B-it} & 53.86 \textsubscript{1.7} & \heatcell{g1}  \textbf{57.96} \textsubscript{1.6} & \heatcell{g1}  \underline{57.47} \textsubscript{3.3} &  54.84 \textsubscript{0.2} \\ 
& \textsc{Qwen3-8B} & \heatcell{g1} 60.43 \textsubscript{1.5} & \heatcell{g5} \textbf{80.95} \textsubscript{1.2} & \heatcell{g4} \underline{79.47} \textsubscript{0.6} & \heatcell{g4} 77.83 \textsubscript{1.2} \\ 
\midrule

\multirow{8}{*}{\rotatebox{90}{\footnotesize \textsc{SPR}}}
& \textsc{GPT-5.1} & \heatcell{g4} \underline{68.67} \textsubscript{0.5} & \heatcell{g4} 66.33 \textsubscript{3.7} & \heatcell{g4} 67.33 \textsubscript{3.7} & \heatcell{g5} \textbf{70.33} \textsubscript{0.5} \\ 
& \textsc{Gemini3-flash} & \heatcell{g7} 89.33 \textsubscript{0.6} & \heatcell{g7} \underline{89.67} \textsubscript{0.5} & \heatcell{g7} 89.33 \textsubscript{1.9} & \heatcell{g7} \textbf{90.33} \textsubscript{0.4} \\ 
\cmidrule{2-6}
& \textsc{Olmo3.1-32B-it} & \heatcell{g1} 43.67 \textsubscript{0.5} & \heatcell{g6} \textbf{81.00} \textsubscript{1.6} & \heatcell{g3} \underline{57.00} \textsubscript{0.8} & \heatcell{g2} 55.33 \textsubscript{0.9} \\ 
& \textsc{Llama3.3-70B-it} & \heatcell{g1} 46.67 \textsubscript{0.5} & \heatcell{g4} \textbf{67.33} \textsubscript{3.1} & \heatcell{g4} 65.67 \textsubscript{4.2} & \heatcell{g2} \underline{67.00} \textsubscript{0.8} \\ 
& \textsc{Qwen3-32B} & \heatcell{g3} \underline{60.67} \textsubscript{0.5} & \heatcell{g2} 59.00 \textsubscript{0.8} & \heatcell{g2} 59.33 \textsubscript{2.9} & \heatcell{g3} \textbf{61.67} \textsubscript{2.9} \\ 
\cmidrule{2-6}
& \textsc{Olmo3-7B-it} & \heatcell{g1} 41.00 \textsubscript{0.8} & \heatcell{g5} \textbf{70.67} \textsubscript{0.9} & \heatcell{g4} \underline{69.00} \textsubscript{3.7} & \heatcell{g4} 68.00 \textsubscript{1.4} \\ 
& \textsc{Llama3.1-8B-it} & 34.67 \textsubscript{2.5} &  \textbf{38.00} \textsubscript{2.8} &  \underline{37.00} \textsubscript{1.6} & 36.00 \textsubscript{2.2} \\ 
& \textsc{Qwen3-8B} & \heatcell{g1} 40.00 \textsubscript{0.0} & \heatcell{g3} \underline{60.67} \textsubscript{0.4} & \heatcell{g2} 58.00 \textsubscript{2.4} & \heatcell{g3} \textbf{61.00} \textsubscript{2.9} \\ 

\midrule

\multirow{8}{*}{\rotatebox{90}{\footnotesize $\alpha NLI$}}
& \textsc{GPT-5.1} & \heatcell{g5} 85.00 \textsubscript{1.6} & \heatcell{g6} 89.00 \textsubscript{0.8} & \heatcell{g7} \textbf{90.00} \textsubscript{0.8} & \heatcell{g7} \textbf{90.00} \textsubscript{0.8} \\ 
& \textsc{Gemini3-flash} & \heatcell{g7} \textbf{90.33}\textsubscript{1.5} & \heatcell{g6} 88.67 \textsubscript{0.8} & \heatcell{g6} 88.33 \textsubscript{0.9} & \heatcell{g7} \textbf{90.33} \textsubscript{1.3} \\ 
\cmidrule{2-6}
& \textsc{Olmo3.1-32B-it} & \heatcell{g4}  75.67 \textsubscript{0.5} &  33.33 \textsubscript{0.9} &  \heatcell{g1} 44.67 \textsubscript{0.5} & \heatcell{g1} \heatcell{g2} 59.33 \textsubscript{0.5} \\ 
& \textsc{Llama3.3-70B-it} & \heatcell{g5}  \underline{83.00} \textsubscript{0.0} & \heatcell{g3} \textbf{83.33} \textsubscript{0.5} & \heatcell{g5} 81.33 \textsubscript{0.9} & \heatcell{g5} \underline{83.00} \textsubscript{0.8} \\ 
& \textsc{Qwen3-32B} & \heatcell{g4} 76.00 \textsubscript{0.0} & \heatcell{g5} \textbf{83.33} \textsubscript{1.2} & \heatcell{g4} 79.67 \textsubscript{0.9} & \heatcell{g5} \underline{81.00} \textsubscript{0.8} \\ 
\cmidrule{2-6}
& \textsc{Olmo3-7B-it} & \heatcell{g2} \textbf{60.00} \textsubscript{0.0} &  25.33 \textsubscript{2.5} &  \heatcell{g1} 48.00 \textsubscript{0.8} & \heatcell{g2} \underline{57.33} \textsubscript{3.1} \\ 
& \textsc{Llama3.1-8B-it} & \heatcell{g3} \underline{64.00} \textsubscript{0.8} & \heatcell{g3} \textbf{67.00} \textsubscript{1.6} & \heatcell{g2} \underline{64.00} \textsubscript{3.6} & \heatcell{g2} 62.67 \textsubscript{2.5} \\ 
& \textsc{Qwen3-8B} & \heatcell{g3} \textbf{69.00} \textsubscript{0.8} & \heatcell{g3} 67.33 \textsubscript{2.5} & \heatcell{g2} 66.33 \textsubscript{3.3} & \heatcell{g2} \textbf{69.00} \textsubscript{0.8} \\ 
\midrule

\multirow{8}{*}{\rotatebox{90}{\footnotesize \textsc{RECV}}}
& \textsc{GPT-5.1} & \heatcell{g6} 83.16 \textsubscript{1.3} & \heatcell{g5} 77.10 \textsubscript{1.3} & \heatcell{g6} \textbf{83.84} \textsubscript{0.8} & \heatcell{g6} \textbf{83.84} \textsubscript{2.5} \\ 
& \textsc{Gemini3-flash} & \heatcell{g6} 81.48 \textsubscript{1.1} & \heatcell{g6} 80.47 \textsubscript{0.8} & \heatcell{g7} \textbf{86.53} \textsubscript{0.6} & \heatcell{g7} \textbf{86.53} \textsubscript{1.9} \\ 
\cmidrule{2-6}
& \textsc{Olmo3.1-32B-it} & \heatcell{g4}  72.39 \textsubscript{0.5} & \heatcell{g5} \underline{79.12} \textsubscript{2.5} & \heatcell{g6} \textbf{80.47} \textsubscript{1.0} & \heatcell{g5} 78.45 \textsubscript{0.5} \\ 
& \textsc{Llama3.3-70B-it} & \heatcell{g5}  79.80 \textsubscript{0.0} & \heatcell{g6} \underline{80.81} \textsubscript{1.4} & \heatcell{g6} \textbf{83.84} \textsubscript{0.8} & \heatcell{g6} 81.14 \textsubscript{1.7} \\ 
& \textsc{Qwen3-32B} & \heatcell{g6} \underline{82.49} \textsubscript{0.5} & \heatcell{g5} 75.76 \textsubscript{0.8} & \heatcell{g5} 78.11 \textsubscript{0.5} & \heatcell{g6} \textbf{84.18} \textsubscript{1.3} \\ 
\cmidrule{2-6}
& \textsc{Olmo3-7B-it} & \heatcell{g3} \textbf{69.36} \textsubscript{1.7} & \heatcell{g3} 67.00 \textsubscript{2.1} & \heatcell{g3} 65.66 \textsubscript{1.6} & \heatcell{g3} \underline{68.01} \textsubscript{2.5} \\ 
& \textsc{Llama3.1-8B-it} & \heatcell{g5}  78.45 \textsubscript{0.5} & \heatcell{g6} 80.47 \textsubscript{3.3} & \heatcell{g6} \textbf{83.84} \textsubscript{0.0} & \heatcell{g6} \underline{83.50} \textsubscript{1.0} \\ 
& \textsc{Qwen3-8B} & \heatcell{g5} \textbf{76.09} \textsubscript{0.5} & \heatcell{g4} 72.05 \textsubscript{1.9} & \heatcell{g4} 74.75 \textsubscript{1.6} & \heatcell{g5} \textbf{76.09} \textsubscript{1.9} \\ 
\midrule
\bottomrule
\end{tabular}%
}
\small\caption{The average accuracies of the final answer and their standard deviations  with respect to the instructed reasoning types.} 
\label{tab:results_wrt_instructed_type}
\end{center}
\end{table}

\section{Complete Accuracy Results}
\label{sec:more_accuracy_results}

Table \ref{tab:sensible_and_compliance_accuracy} shows the numeric accuracies of the judge-inferred $S \cap C$, $S \cap \neg C$, and $\neg S \cap C$ categories in the four datasets, which is corresponding to Figure \ref{fig:accuracy_wrt_sensible_compliance_category}.
Table \ref{tab:results_wrt_instructed_type} shows the accuracy results grouped by the instructed reasoning type $\mathbf{t'}$.

\section{Steering Results}
\label{app:more_steering}
\begin{figure*}[ht]
     \centering
     \begin{subfigure}[b]{0.32\textwidth}
         \centering
         \includegraphics[width=\textwidth]{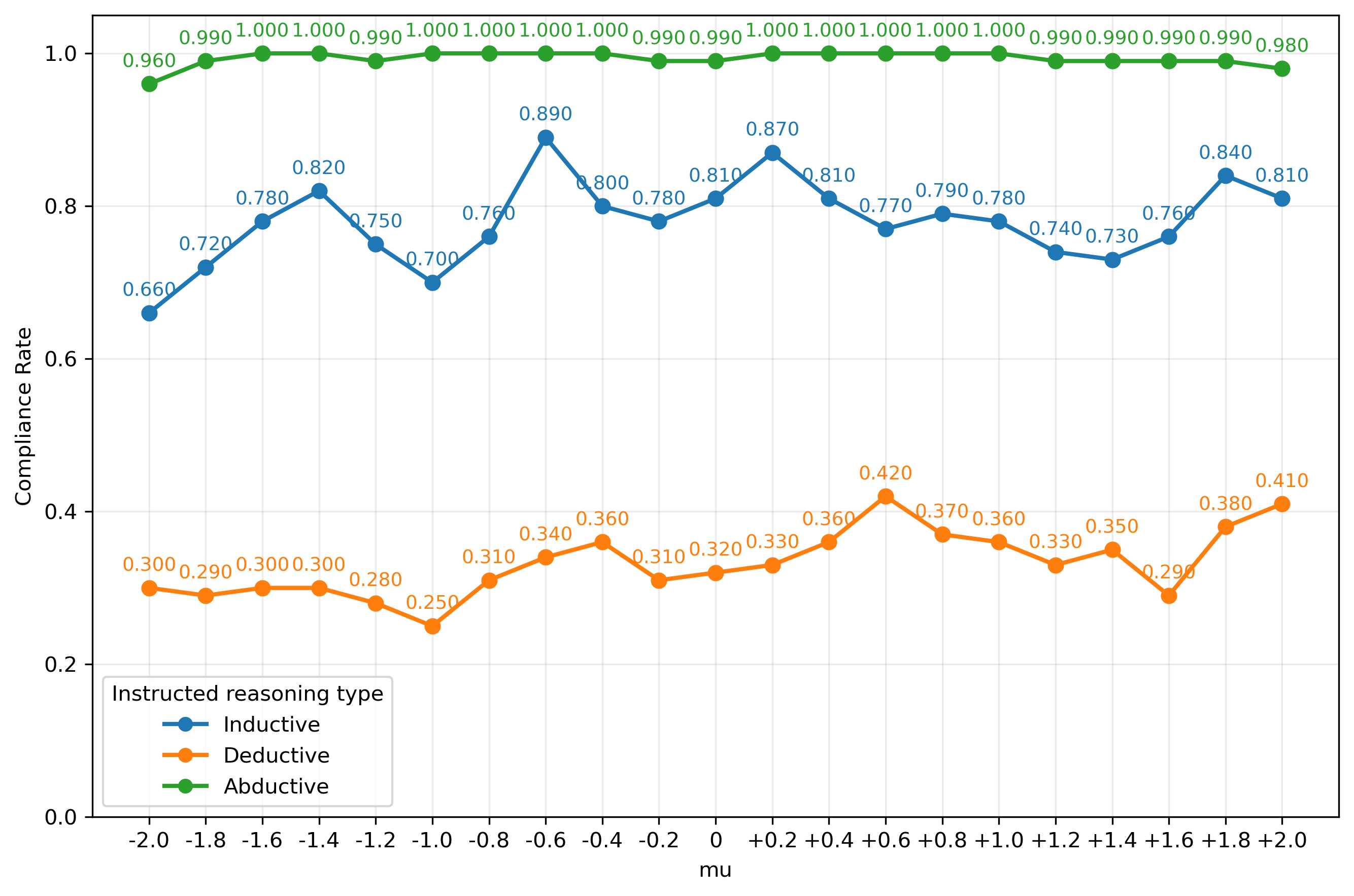}
         \caption{Compliance Rate}
         \label{fig:compliance_steer_llama}
     \end{subfigure}
     \hfill
     \begin{subfigure}[b]{0.32\textwidth}
         \centering
         \includegraphics[width=\textwidth]{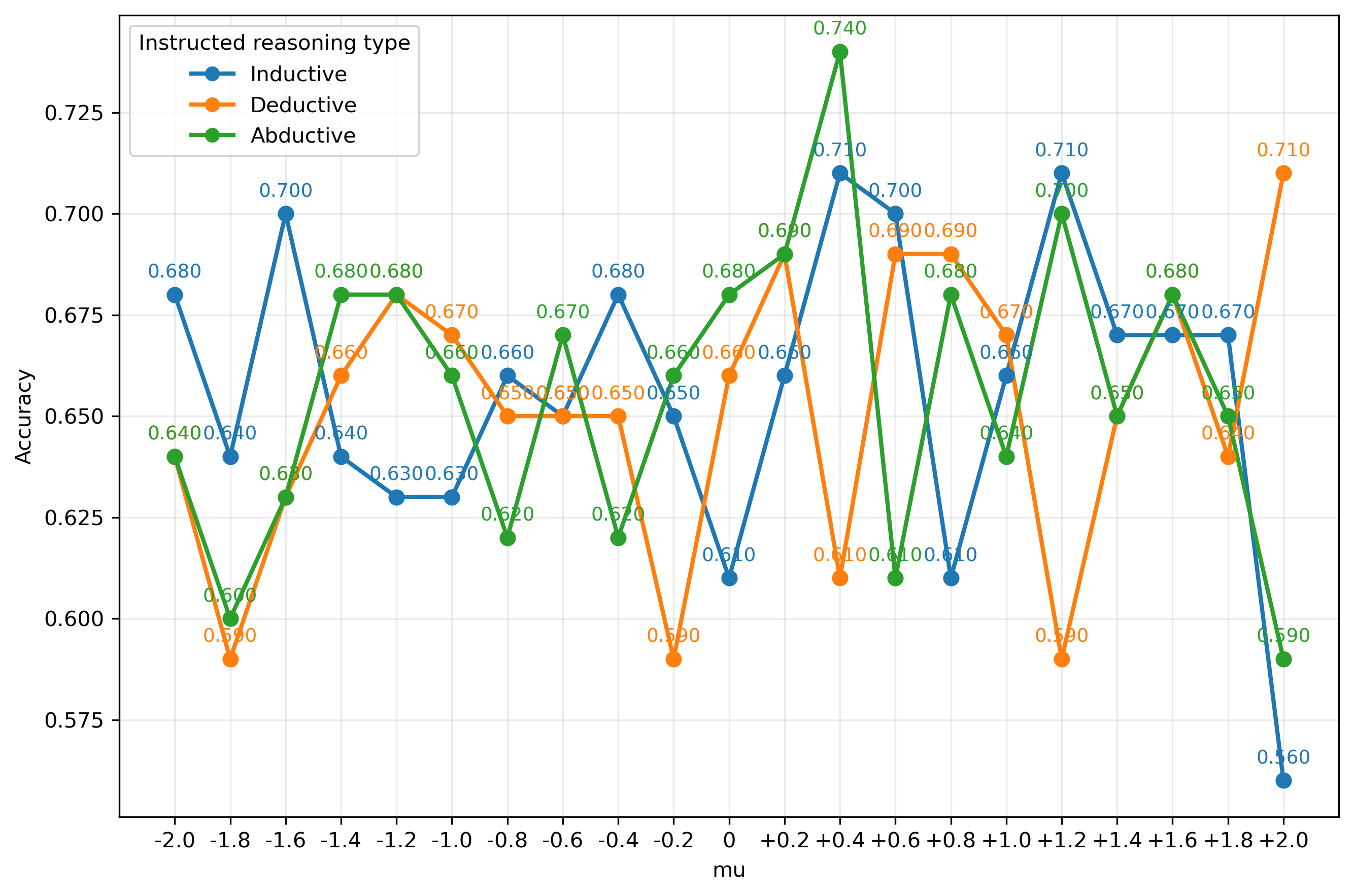}
         \caption{Final Answer Accuracy}
         \label{fig:accuracy_steer_llama}
     \end{subfigure}
     \hfill
     \begin{subfigure}[b]{0.32\textwidth}
         \centering
         \includegraphics[width=\textwidth]{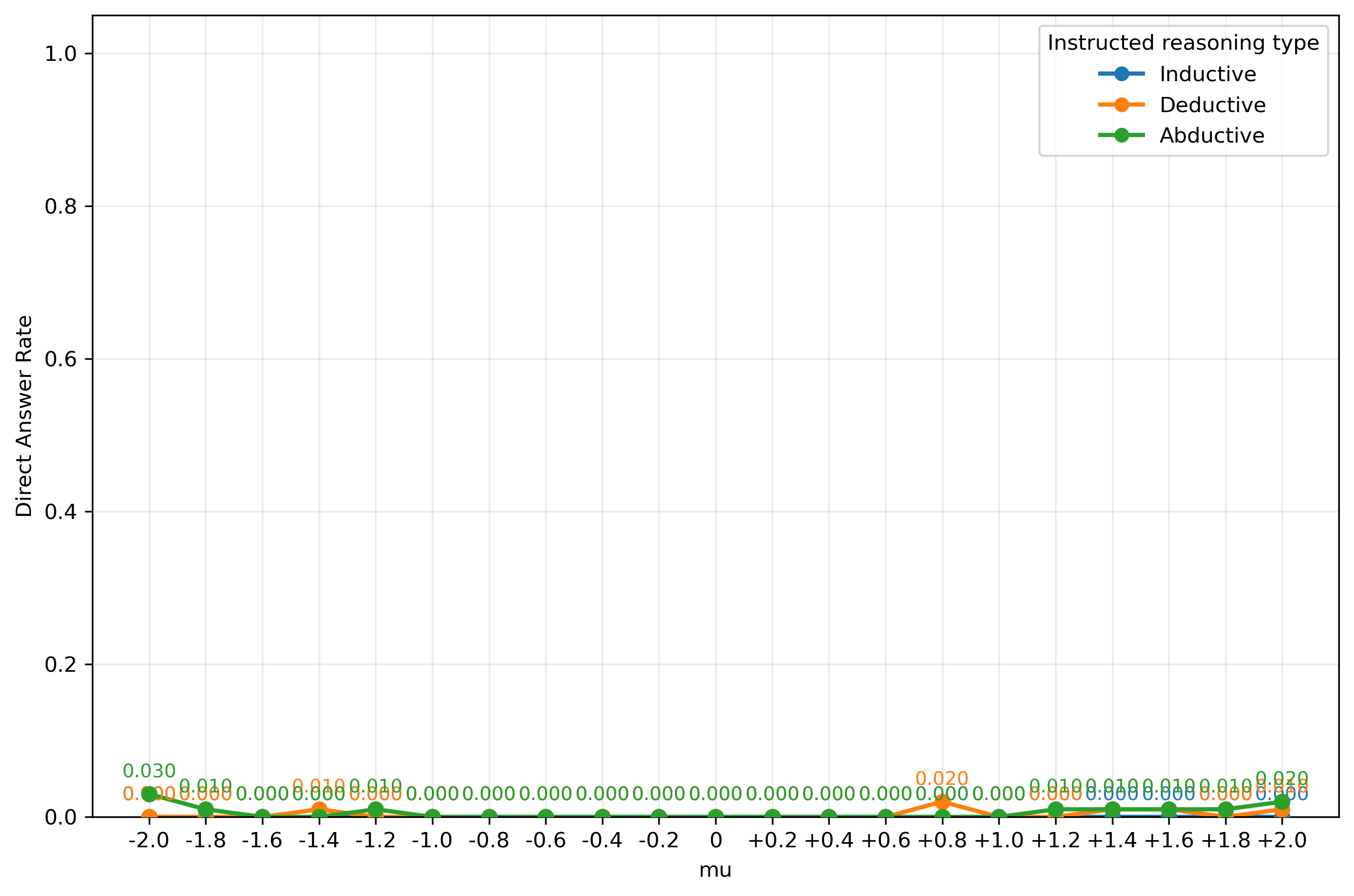}
         \caption{Direct Answer Ratio}
         \label{fig:direct_steer_llama}
     \end{subfigure}
        \caption{The impact of multiplier $\mu$ across reasoning types on $\alpha$-NLI when steering the layer 14-17 of \textsc{Llama3.1-8B-IT}.}
        \label{fig:steering_llama_anli}
\end{figure*}

Figure \ref{fig:steering_llama_anli} show the steering results of \textsc{Llama3.1-8B-IT} on $\alpha$-NLI. The compliance rates are moderately increased with higher $\mu$ values.

\end{document}